\documentclass[11pt]{article}

\usepackage[preprint]{acl}

\usepackage{times}
\usepackage{latexsym}

\usepackage[T1]{fontenc}
\usepackage[utf8]{inputenc}

\usepackage{microtype}

\usepackage{inconsolata}

\usepackage{enumitem}
\usepackage{subcaption}
\usepackage{graphicx}
\usepackage[table]{xcolor}
\usepackage{multirow}

\usepackage{booktabs}

\usepackage{amsmath,amssymb}
\usepackage{amsthm}
\usepackage{amsfonts}

\theoremstyle{plain}

\theoremstyle{definition}

\newtheorem{assumption}{Assumption}

\usepackage[most]{tcolorbox}
\usepackage{fontawesome5}    
\usepackage{enumitem}

\definecolor{prompt-bg}{RGB}{245, 247, 250}
\definecolor{prompt-frame}{RGB}{50, 60, 80}
\definecolor{think-text}{RGB}{80, 80, 80}
\definecolor{think-bg}{RGB}{252, 252, 252}

\definecolor{traj1-frame}{RGB}{70, 130, 180}
\definecolor{traj1-bg}{RGB}{240, 248, 255}

\definecolor{traj2-frame}{RGB}{60, 179, 113}
\definecolor{traj2-bg}{RGB}{245, 255, 250}

\definecolor{traj-red-frame}{RGB}{178, 34, 34} 
\definecolor{traj-red-bg}{RGB}{255, 245, 245}

\definecolor{traj-green-frame}{RGB}{34, 139, 34}
\definecolor{traj-green-bg}{RGB}{240, 255, 240}

\definecolor{maincolor}{RGB}{50, 120, 200}    
\definecolor{teacherAcolor}{RGB}{235, 245, 255} 
\definecolor{teacherBcolor}{RGB}{245, 235, 255} 
\definecolor{qwenpurple}{RGB}{120, 50, 180}

\newtcolorbox{promptbox}{
    enhanced,
    breakable,
    colback=prompt-bg,
    colframe=prompt-frame,
    fonttitle=\bfseries\large,
    title={\ifdefined\faUser\faUser\ \ \fi Prompt / Input},
    attach boxed title to top left={xshift=5mm, yshift*=-3mm},
    boxed title style={colframe=prompt-frame, colback=prompt-frame},
    sharp corners=downhill,
    arc=3mm,
    boxrule=1.2pt,
    drop shadow,     
    top=8mm, bottom=5mm, left=5mm, right=5mm,
    before skip=1em, after skip=1em
}

\newtcolorbox{trajbox}[2][]{
    enhanced,
    breakable,
    colback=#2-bg,
    colframe=#2-frame,
    fonttitle=\bfseries,
    title={#1},
    boxrule=1pt,
    arc=2mm,
    drop shadow,
    left=4mm, right=4mm, top=4mm, bottom=4mm,
    before skip=1em, after skip=1em
}

\newtcolorbox{thinkbox}{
    enhanced,
    breakable,
    frame hidden,
    colback=think-bg,
    borderline west={3pt}{0pt}{gray!30},
    fontupper=\small\color{think-text},
    fonttitle=\bfseries\small\color{gray!80},
    title={$\hookrightarrow$ \textit{Internal Reasoning}},
    attach boxed title to top left={yshift=-2mm, xshift=0mm},
    boxed title style={boxrule=0pt, colframe=white, colback=white},
    coltitle=think-text,
    top=5pt, bottom=5pt, left=10pt, right=5pt,
    parbox=false
}

\title{Bridging SFT and RL: Dynamic Policy Optimization for Robust Reasoning}

\author{
 \textbf{Taojie Zhu}\textsuperscript{1,2,$\dagger$}
 \thanks{Work done during internship at ZTE.},
 \textbf{Dongyang Xu}\textsuperscript{2,$\dagger$,$\ddagger$},
 \textbf{Ding Zou}\textsuperscript{2,$\spadesuit$},
 \textbf{Sen Zhao}\textsuperscript{3},
 \textbf{Qiaobo Hao}\textsuperscript{2},
 \textbf{Zhiguo Yang}\textsuperscript{2},
 \textbf{Yonghong He}\textsuperscript{1,$\ddagger$}
\\
\\
 \textsuperscript{1}Shenzhen International Graduate School, Tsinghua University
\\
 \textsuperscript{2}Intelligent System Department, Zhongxing Telecom Equipment (ZTE)
\\
 \textsuperscript{3}Institute of Advanced Interdisciplinary Studies, Chongqing University of Posts and Telecommunications
\\
\\
 \textsuperscript{$\dagger$}Equal contribution. \quad
 \textsuperscript{$\ddagger$}Corresponding author. \quad
 \textsuperscript{$\spadesuit$}Project Leader.
\\
 \texttt{xu.dongyang2@zte.com.cn}, \texttt{heyh@sz.tsinghua.edu.cn}
}

\begin{document}
\maketitle

\begin{abstract}
Post-training paradigms for Large Language Models (LLMs), primarily Supervised Fine-Tuning (SFT) and Reinforcement Learning (RL), face a fundamental dilemma: SFT provides stability (low variance) but suffers from high fitting bias, while RL enables exploration (low bias) but grapples with high gradient variance. Existing unified optimization strategies often employ naive loss weighting, overlooking the statistical conflict between these distinct gradient signals. In this paper, we provide a rigorous theoretical analysis of this bias-variance trade-off and propose \textbf{DYPO} (Dynamic Policy Optimization), a unified framework designed to structurally mitigate this conflict. DYPO integrates three core components: (1) a \textit{Group Alignment Loss (GAL)} that leverages intrinsic group dynamics to significantly reduce RL gradient variance; (2) a \textit{Multi-Teacher Distillation} mechanism that corrects SFT fitting bias via diverse reasoning paths; and (3) a \textit{Dynamic Exploitation-Exploration Gating} mechanism that adaptively arbitrates between stable SFT and exploratory RL based on reward feedback. Theoretical analysis confirms that DYPO linearly reduces fitting bias and minimizes overall variance. Extensive experiments demonstrate that DYPO significantly outperforms traditional sequential pipelines, achieving an average improvement of 4.8\% on complex reasoning benchmarks and 13.3\% on out-of-distribution tasks. Our code is publicly available at \url{https://github.com/Tocci-Zhu/DYPO}.
\end{abstract}

\section{Introduction}

The reasoning capabilities of Large Language Models (LLMs) have become a central focus in artificial intelligence~\cite{jaech2024openai,guo2025deepseek,team2025kimi}. While reasoning-guidance techniques like Chain-of-Thought (CoT) prompting have significantly advanced model performance on multi-step tasks~\cite{wei2022chain}, traditional prompting methods relying on static templates struggle with scalability and dynamic adaptability. Consequently, the research focus has shifted toward the post-training stage to enhance robustness and generalization~\cite{wang2023aligninglargelanguagemodels}.
\begin{figure*}[htbp]
    \centering

    \begin{subfigure}[b]{0.45\linewidth}
        \centering
        \includegraphics[width=\linewidth]{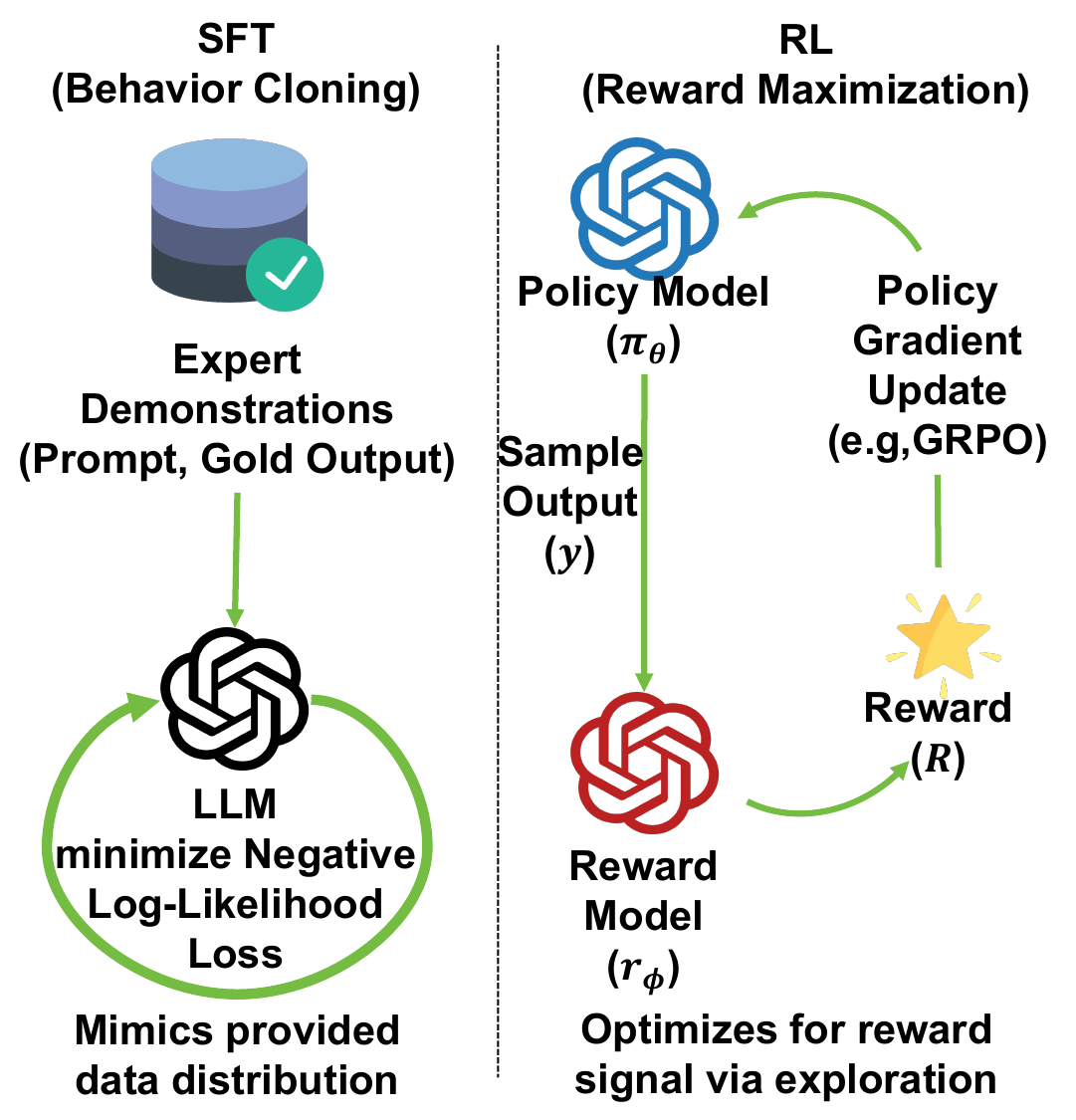}
        \caption{Data Flow \& Training Mechanisms}
        \label{fig:fig_a}
    \end{subfigure}
    \hfill
    \begin{minipage}[b]{0.53\linewidth}

        \begin{subfigure}[b]{\linewidth}
            \centering
            \includegraphics[width=\linewidth]{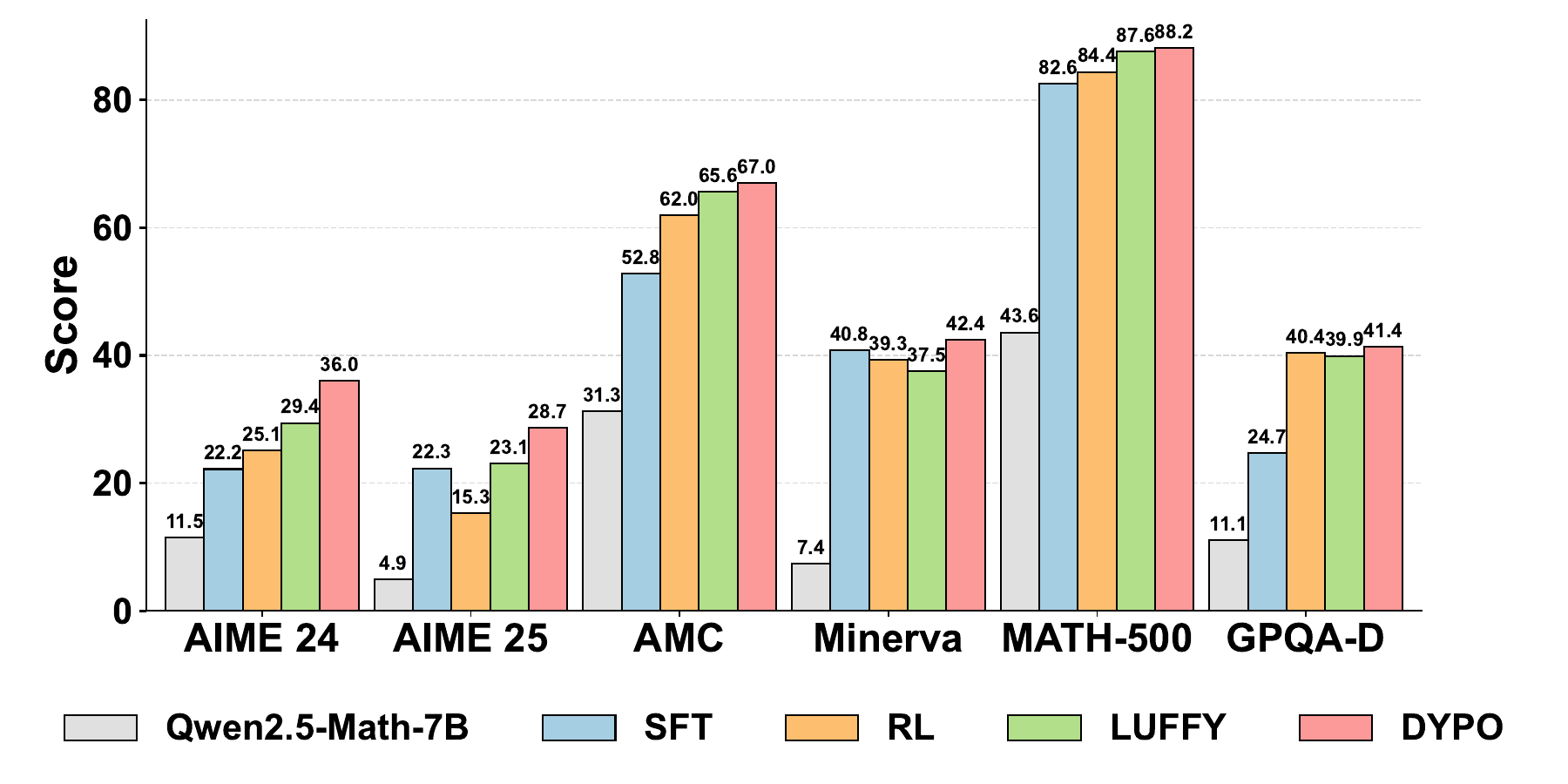}
            \vspace{-20pt}
            \caption{Model performance}
            \label{fig:fig_b}
        \end{subfigure}
        
        \vspace{0pt}

        \begin{subfigure}[b]{\linewidth}
            \centering
            \includegraphics[width=\linewidth]{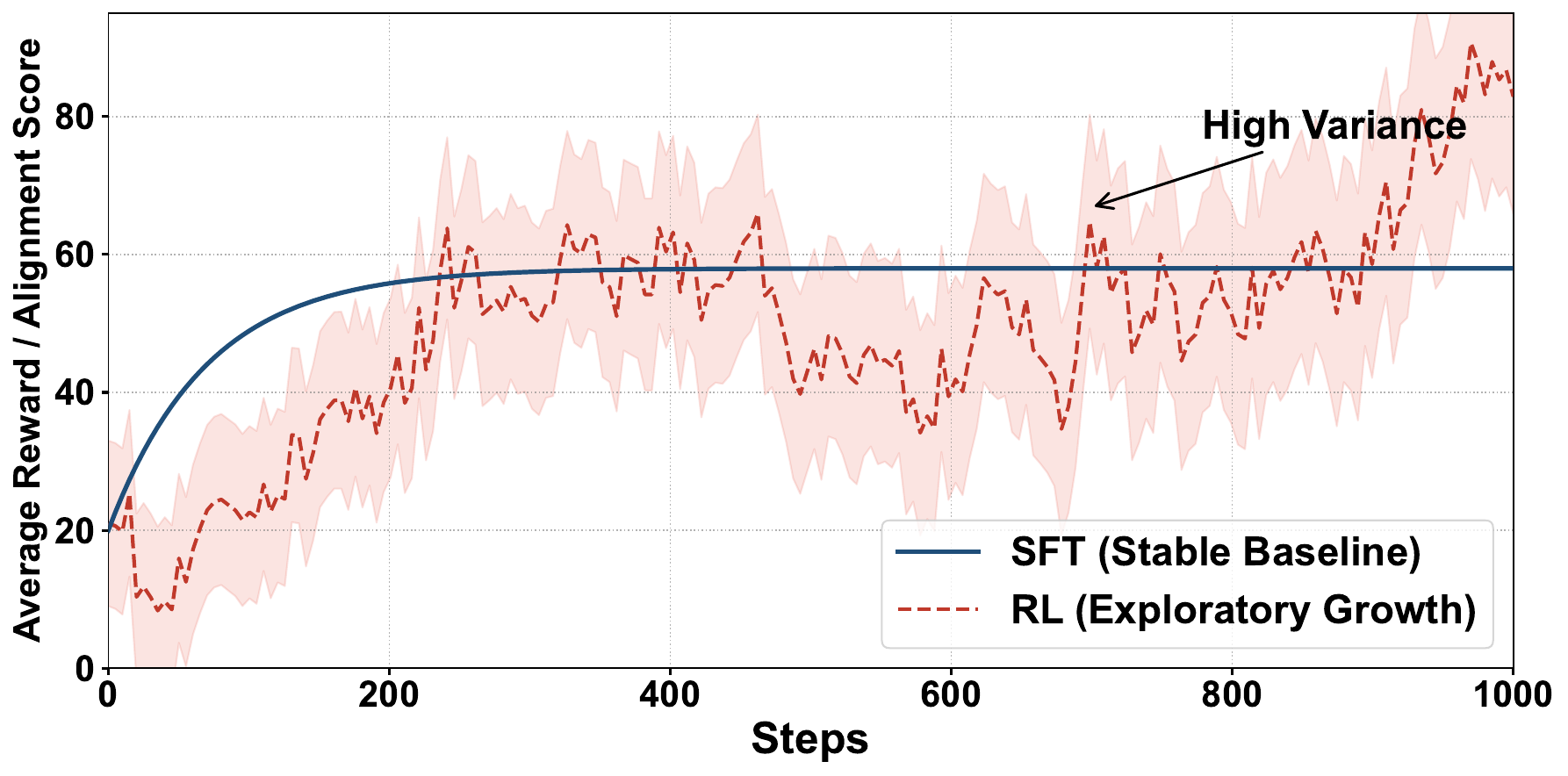}
            \vspace{-20pt}
            \caption{Learning Curves}
            \label{fig:fig_c}
        \end{subfigure}
    \end{minipage}
    \vspace{0pt}
    \caption{\textbf{The SFT-RL Dilemma:} Balancing the high-bias stability of SFT against the high-variance exploration of RL.}
    \label{fig:sft_rl_comparison}
\end{figure*}
Current mainstream post-training paradigms generally fall into two categories:
i) Supervised Fine-Tuning (SFT): SFT offers efficient knowledge injection by learning from high-quality CoT corpora~\cite{sanh2022multitask,wei2021finetuned}. Its low-variance nature ensures stability and rapid fitting, but often at the cost of limited exploratory capacity and restricted Out-of-Distribution (OOD) generalization.
ii) Reinforcement Learning (RL): Methods such as RLHF or RLVR allow models to autonomously explore the reasoning space via reward signals, substantially enhancing generalization~\cite{ouyang2022training,ramamurthy2022reinforcement,ppo,grpo}. Unlike SFT, RL relies on the base model's intrinsic capabilities; consequently, weaker models often struggle to capture sparse reward signals in complex tasks.

To combine these strengths, researchers have widely adopted a ``SFT-then-RL'' training pipeline~\cite{touvron2023llama,yoshihara2025practical}. However, this sequential approach suffers from \textit{bias propagation}, where SFT-induced biases misguide subsequent RL exploration~\cite{lv2025towards}, alongside significant computational overhead.

Recent research has therefore shifted toward \textit{unified optimization}, which combines SFT and RL objectives within a single training process~\cite{yan2025learningreasonoffpolicyguidance,fu2025srft,zhang2025onpolicyrlmeetsoffpolicy,chen2025beyond}. Representative methods include SuperRL~\cite{liu2025superrlreinforcementlearningsupervision}, which adopts a binary switching strategy between supervision and reinforcement learning, and CHORD~\cite{zhang2025onpolicyrlmeetsoffpolicy}, which harmonizes the two objectives through dynamic soft weighting. These approaches highlight the growing interest in unified SFT-RL post-training, but they still apply a largely uniform optimization recipe across samples whose learning signals differ fundamentally in reliability.

Despite the growing interest in unified SFT-RL training, existing fusion strategies predominantly operate at a ``surface level'' via simple loss weighting~\cite{lv2025towards}. This approach overlooks two fundamental issues. First, it ignores the inherent statistical conflict between the gradient signals: SFT gradients are \textit{high-bias} (fitting static data) but \textit{low-variance}~\cite{wu2025generalizationofsft}, whereas RL gradients are \textit{low-bias} (reward-driven) but \textit{high-variance} (due to sampling stochasticity)~\cite{ramamurthy2022reinforcement}. Naively aggregating these conflicting vectors is sub-optimal, as RL's high variance destabilizes training while SFT's high bias constrains exploration. Second, this uniform approach fails to account for regime-dependent differences in sample difficulty. Specifically, trivial samples provide marginal optimization signals since model performance is already saturated; hard samples yield extremely sparse rewards, rendering RL highly inefficient; and only mid-difficulty samples simultaneously preserve reward discrimination and expose meaningful failure modes. Consequently, globally mixing SFT and RL objectives cannot fully resolve the multidimensional mismatch between stable but biased supervision and exploratory but high-variance policy optimization.

In this paper, we first provide a theoretical analysis formally defining this bias-variance trade-off in SFT-RL fusion. We then propose \textbf{DYPO} (\underline{D}\underline{Y}namic \underline{P}olicy \underline{O}ptimization), a unified framework that introduces structural solutions to concurrently mitigate both limitations.Unlike binary switching or soft weighting methods, DYPO performs instance-level routing based on rollout outcomes and assigns each regime to a distinct optimization objective.

Specifically, DYPO comprises three core components:
% [leftmargin=*]
\begin{itemize}[leftmargin=*]
    \item \textbf{Dynamic Difficulty Grading}: A mechanism that dynamically categorizes queries based on group rollout outcomes. It effectively arbitrates the optimization pathway: routing complete failures (Hard) to stable SFT for knowledge injection, while directing inconsistent attempts (Mid) to low-bias RL for exploration.
    \item \textbf{Bias Correction (SFT)}: For `Hard' samples, we employ a Multi-Teacher Distillation mechanism to correct the fitting bias inherent in SFT by aggregating diverse reasoning paths from different teacher models.
    \item \textbf{Variance Reduction (RL)}: For `Mid' samples, we introduce a Group Alignment Loss (GAL)~\cite{rafailov2023direct} that leverages intrinsic group dynamics.By effectively reinforcing winning samples while suppressing losing ones, GAL significantly reduces RL gradient variance compared to standard pairwise losses.
\end{itemize}

Theoretically, we prove that our {Dynamic Difficulty Grading} mechanism minimizes overall variance by strategically allocating queries based on reward feedback. For `Hard' samples, the triggered {multi-teacher strategy} linearly reduces fitting bias; for `Mid' samples, the {GAL} reduces gradient variance by orders of magnitude compared to GRPO. Experimentally, \textsc{DYPO} yields 5--10\% performance gains on complex reasoning benchmarks.

\section{Preliminaries}
\label{sec:preliminaries}

In this section, we formalize the reasoning trace generation problem and review the two foundational paradigms: SFT and RL. We specifically highlight their respective statistical challenges—fitting bias in SFT and gradient variance in RL—which motivate our proposed approach.

\subsection{Problem Formulation}
We model the reasoning task as a sequential decision-making process. Given an input prompt $q$ sampled from a distribution $\mathcal{D}$, the LLM functions as a stochastic policy $\pi_{\theta}(\tau|q)$ parameterized by $\theta$. Here, $\tau = (a_1, a_2, \dots, a_T)$ represents a reasoning trajectory consisting of a sequence of tokens. The probability of generating a trajectory is factorized autoregressively:
\begin{equation}
    \pi_{\theta}(\tau|q) = \prod_{t=1}^{T} \pi_{\theta}(a_t | q, a_{<t})
\end{equation}
Upon completion, the trajectory $\tau$ is evaluated by a reward function $R(q, \tau) \in \mathbb{R}$. The objective is to maximize the expected reward $J(\theta) = \mathbb{E}_{q \sim \mathcal{D}} \mathbb{E}_{\tau \sim \pi_{\theta}(\cdot|q)} [R(q, \tau)]$\cite{ouyang2022training,ziegler2019fine}.

\subsection{SFT and Fitting Bias}
Standard SFT adapts the policy by minimizing the negative log-likelihood on a static dataset $\mathcal{D}_{\text{sft}}$ containing gold-standard pairs $(q, \tau^*)$\cite{touvron2023llama,wei2021finetuned}:
\begin{equation}
    \mathcal{L}_{\text{SFT}}(\theta) = \mathbb{E}_{(q, \tau^*) \sim \mathcal{D}_{\text{sft}}} \left[ - \log \pi_{\theta}(\tau^* | q) \right]
\end{equation}
While SFT provides stable supervision, it inherently suffers from \textbf{fitting bias}. Since the optimization is constrained to the fixed support of $\mathcal{D}_{\text{sft}}$, the model tends to overfit the specific distribution of the single teacher or dataset. This mimicry limits the model's ability to explore novel reasoning paths and often leads to sub-optimal local minima where the policy fails to generalize beyond the training examples.

\subsection{GRPO and Gradient Variance}
To enable exploration, we employ GRPO\cite{grpo}. For each prompt $q$, GRPO samples a group of trajectories $G = \{\tau_1, \tau_2, \dots, \tau_k\}$ and optimizes the policy using group-normalized advantages:
\begin{equation}
\begin{array}{r@{\,}l}
    \mathcal{L}_{\text{GRPO}}(\theta) &= -\mathbb{E}_{q \sim \mathcal{D}} \Bigg[ \frac{1}{k} \sum_{i=1}^{k} \Big( \text{CLIP}(\rho_i, \hat{A}_i, \epsilon) \Big) \\[-4pt] 
    & \mspace{25mu} - \beta_{\text{KL}} \mathbb{D}_{\text{KL}}(\pi_\theta || \pi_{\text{ref}}) \Bigg]
\end{array}
\end{equation}
where $\rho_i$ is the probability ratio and $\hat{A}_i$ is the advantage computed by standardizing rewards within the group $G$. Although GRPO provides an low-biased objective for reward maximization, it introduces high gradient variance. This instability arises from the stochastic nature of trajectory sampling and the reliance on a sparse reward signal. With a limited group size $k$, the Monte Carlo estimate of the gradient can be highly noisy, often destabilizing the training process in complex reasoning tasks.

\section{Methodology}
\label{sec:method}

\begin{figure*}[t]
    \centering
    \includegraphics[width=\linewidth]{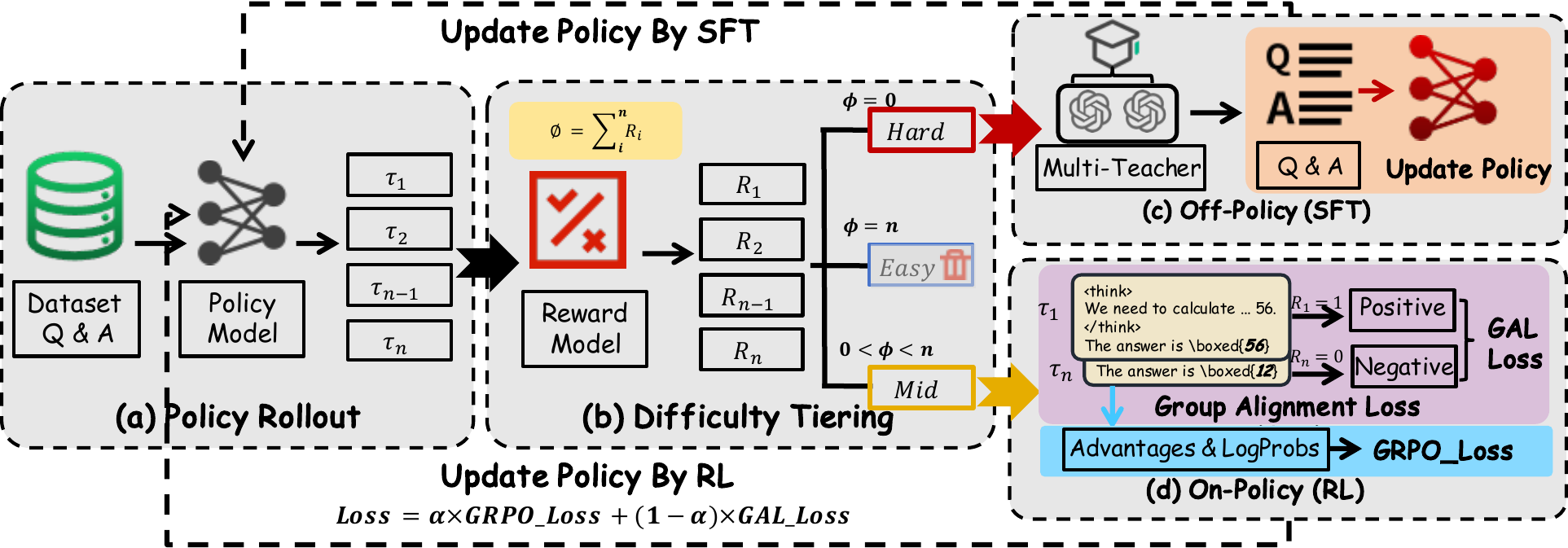}
    \caption{The overall framework of DYPO. The system employs a Dynamic Difficulty Grading mechanism to categorize queries into Easy, Hard, and Mid tiers based on group rollout outcomes, dispatching them to the most effective optimization pathway.}
    \label{fig:workflow}
\end{figure*}

In this section, we present \textbf{DYPO}, a unified framework that dynamically balances exploration and stability by routing queries to the most suitable optimization pathway. The key intuition is that different queries expose learning signals of different reliability: easy queries are already saturated, hard queries lack usable reward signals, and only mid-difficulty queries preserve informative relative feedback for RL.

Formally, the unified objective of DYPO is constructed as a dynamic mixture of supervised and reinforcement learning objectives.
\begin{equation}
\label{eq:dypo_total}
\begin{array}{r@{\,}l}
    \mathcal{L}_{\text{DYPO}}(\theta) &= \mathbb{E}_{q} \bigg[ 
    \underbrace{\mathbb{I}_{\mathcal{H}}(q) \cdot \gamma \mathcal{L}_{\text{SFT}}}_{\text{Bias Mitigation (Sec.~\ref{sec:method_hard})}} \\[0pt] 
    & \mspace{6mu} + \underbrace{\mathbb{I}_{\mathcal{M}}(q) \cdot \big( \alpha \mathcal{L}_{\text{GRPO}} + (1-\alpha) \mathcal{L}_{\text{GAL}} \big)}_{\text{Variance Reduction (Sec.~\ref{sec:method_mid})}} 
    \bigg]
\end{array}
\end{equation}

where $\mathbb{I}_{\mathcal{H}}$ and $\mathbb{I}_{\mathcal{M}}$ are indicator functions determined by a difficulty grading mechanism (Sec.~\ref{sec:method_unified}). Easy samples have zero contribution to the training objective and are therefore omitted from Eq.~\eqref{eq:dypo_total} for simplicity. The coefficients $\gamma$ and $\alpha$ control the strength of distillation and the bias-variance trade-off in RL, respectively. Through the structural separation of the learning process, Eq.~\eqref{eq:dypo_total} allows DYPO to better manage the bias-variance trade-off across different learning stages.

\subsection{Dynamic Difficulty Grading}
\label{sec:method_unified}
We propose a strategy to distinguish data samples based on their impact on training variance, termed Dynamic Difficulty Grading, to optimize the bias-variance trade-off. Specifically, we aim to filter out trivial instances yielding negligible gradients and overly complex outliers that induce high variance, thereby isolating the informative samples most conducive to robust optimization.

Specifically, given a query $q$, the policy $\pi_{\theta}$ generates a group of $k$ trajectories $G = \{\tau_1, \dots, \tau_k\}$. Let $R(\tau_i) \in \{0, 1\}$ denote the binary correctness reward. We categorize the training instance into three levels based on the reward distribution:
\begin{itemize}[leftmargin=*, topsep=2pt, itemsep=0pt, parsep=0pt]
    \item \textbf{Easy ($\mathcal{E}$):} The model solves the problem consistently ($\forall \tau \in G, R(\tau)=1$). These samples provide diminishing returns for gradient estimation and are \textbf{discarded} for efficiency.
    \item \textbf{Hard ($\mathcal{H}$):} The model fails completely ($\forall \tau \in G, R(\tau)=0$). In this regime, valid reward signals are unavailable, causing standard RL gradients to fail. To bridge this gap, we adopt \textbf{Multi-Teacher Distillation}.
    \item \textbf{Mid ($\mathcal{M}$):} The group contains mixed results ($\exists \tau_i, \tau_j \in G, R(\tau_i) \neq R(\tau_j)$). This represents the critical learning frontier. We apply a hybrid objective of \textbf{GRPO} and \textbf{GAL} to leverage the relative feedback.
\end{itemize}

By accurately categorizing each sample into distinct difficulty levels, we integrate a refined sample stratification into our unified optimization framework. This ensures the model prioritizes the most effective learning signals during training. Subsequently, we detail how this framework leverages such stratification to effectively balance variance and bias.

\subsection{Mitigating Supervisory Bias via Multi-Teacher Distillation}
\label{sec:method_hard}
For instances falling into the Hard regime ($\mathbb{I}_{\mathcal{H}}=1$), the model suffers from insufficient prior knowledge to formulate valid reasoning paths, making autonomous exploration prone to failure. To resolve the issue while mitigating the supervisory bias typically associated with single-source supervision, we introduce a \textbf{Multi-Teacher Distillation} strategy.

Rather than relying on a deterministic target from a single source, we maintain an ensemble of $m$ teacher oracles. For each hard query, we uniformly sample a target trajectory $\tau_{\text{tgt}}$ from the candidate set $\{\tau^{(1)}, \dots, \tau^{(m)}\}$ derived from these teachers:
\begin{equation}
\label{eq:sft_multi}
\begin{array}{r@{\,}l}
    \mathcal{L}_{\text{SFT}}(\theta) = \mathbb{E}_{\tau_{\text{tgt}} \sim \mathcal{U}(\{\tau^{(1)}, \dots, \tau^{(m)}\})} \left[ -\log \pi_{\theta}(\tau_{\text{tgt}}|q) \right]
\end{array}
\end{equation}
The theoretical validation for utilizing multiple teachers lies in the decomposition of supervisory bias. Let $\tau^*$ denote the ground-truth optimal reasoning path. A single teacher $i$ provides a supervision signal $\tau^{(i)}$ which deviates from the truth according to the following decomposition:
\begin{equation}
    \tau^{(i)} = \tau^* + \mathbf{b}_{\text{sys}} + \mathbf{b}_i
\end{equation}
Here, $\mathbf{b}_{\text{sys}}$ represents the systematic bias common to all LLMs (e.g., limitation of language modality), while $\mathbf{b}_i$ represents the \textit{idiosyncratic bias} specific to the $i$-th teacher model (e.g., preference for specific formatting or distinct hallucination patterns).

When relying on a single teacher ($m=1$), the student model blindly inherits the full bias vector $\|\mathbf{b}_{\text{sys}} + \mathbf{b}_i\|$. However, under the \textit{diversity assumption}—where different teachers exhibit uncorrelated bias directions (i.e., $\mathbb{E}[\mathbf{b}_i] \approx 0$)—the aggregation of $m$ teachers significantly attenuates the idiosyncratic component. The effective bias of the multi-teacher ensemble is derived by averaging the individual error vectors:
\begin{equation}
\begin{aligned}
    \| \text{Bias}_{\text{multi}} \|^2 &= \left\| \mathbf{b}_{\text{sys}} + \frac{1}{m} \sum_{i=1}^{m} \mathbf{b}_i \right\|^2 \\
    &= \| \mathbf{b}_{\text{sys}} \|^2 + \underbrace{\left\| \frac{1}{m} \sum_{i=1}^{m} \mathbf{b}_i \right\|^2}_{\text{Idiosyncratic Term}}
\end{aligned}
\end{equation}
Assuming independence between teacher biases, the magnitude of the idiosyncratic bias reduces linearly with $m$:
\begin{equation}
    \mathbb{E} \left[ \left\| \frac{1}{m} \sum_{i=1}^{m} \mathbf{b}_i \right\|^2 \right] = \frac{1}{m} \bar{\sigma}_{\text{bias}}^2
\end{equation}
Consequently, we can formally establish that the multi-teacher objective strictly reduces the total supervisory bias compared to the single-teacher baseline ($m=1$):
\begin{align}
    \mathbb{E}[\|\text{Bias}_{\text{multi}}\|^2] &= \|\mathbf{b}_{\text{sys}}\|^2 + \frac{\bar{\sigma}_{\text{bias}}^2}{m} \\
    \mathbb{E}[\|\text{Bias}_{\text{single}}\|^2] &= \|\mathbf{b}_{\text{sys}}\|^2 + \bar{\sigma}_{\text{bias}}^2 \\
    \mathbb{E}[\|\text{Bias}_{\text{multi}}\|^2] &< \mathbb{E}[\|\text{Bias}_{\text{single}}\|^2]
\end{align}
In essence, aggregating supervision signals cancels out the idiosyncratic biases inherent to individual teachers, guiding the model toward the robust intersection of valid reasoning paths. By providing a stabilized policy prior with reduced bias, Multi-Teacher SFT enables effective exploration of the solution space, seamlessly bridging the gap between supervised likelihood maximization and expected reward maximization.

\subsection{Variance-Reduced RL with Group Alignment}
\label{sec:method_mid}

The Mid regime ($\mathbb{I}_{\mathcal{M}}=1$) represents the critical learning frontier where the model exhibits capability but lacks consistency, producing a mixture of correct and incorrect responses. We identify this regime as the target scenario for RL intervention. While Reinforcement Learning is theoretically ideal for amplifying these correct signals to improve performance, standard RL algorithms often struggle with high variance in gradient estimates, which can severely impede convergence stability and speed. To address this bottleneck, we propose a novel optimization strategy: \textbf{Variance-Reduced RL with Group Alignment}.

\paragraph{Instability of GRPO Gradient.}
To motivate our approach, we first examine the gradient of GRPO. For a group of size $k$, the gradient is:
\begin{equation}
    g_{\text{GRPO}} = \frac{1}{k} \sum_{i=1}^k \hat{A}_i \cdot \nabla_\theta \log \pi_\theta(\tau_i|q)
\end{equation}
Let $\Sigma_s \triangleq \mathbb{E}[\|\nabla_\theta \log \pi_\theta\|^2]$ be the variance of the score function. Assuming normalized advantages ($\mathbb{E}[\hat{A}^2] \approx 1$), the variance of this estimator scales as:
\begin{equation}
    \text{Var}(g_{\text{GRPO}}) \approx \frac{1}{k} \Sigma_s
\end{equation}
While increasing $k$ reduces variance, the unbounded nature of $\hat{A}_i$ induces high variance, leading to unstable updates during early exploration.

\begin{figure}[t]
  \centering
  \includegraphics[width=\linewidth, trim={1.8in 1.6in 1.2in 1.8in}, clip]{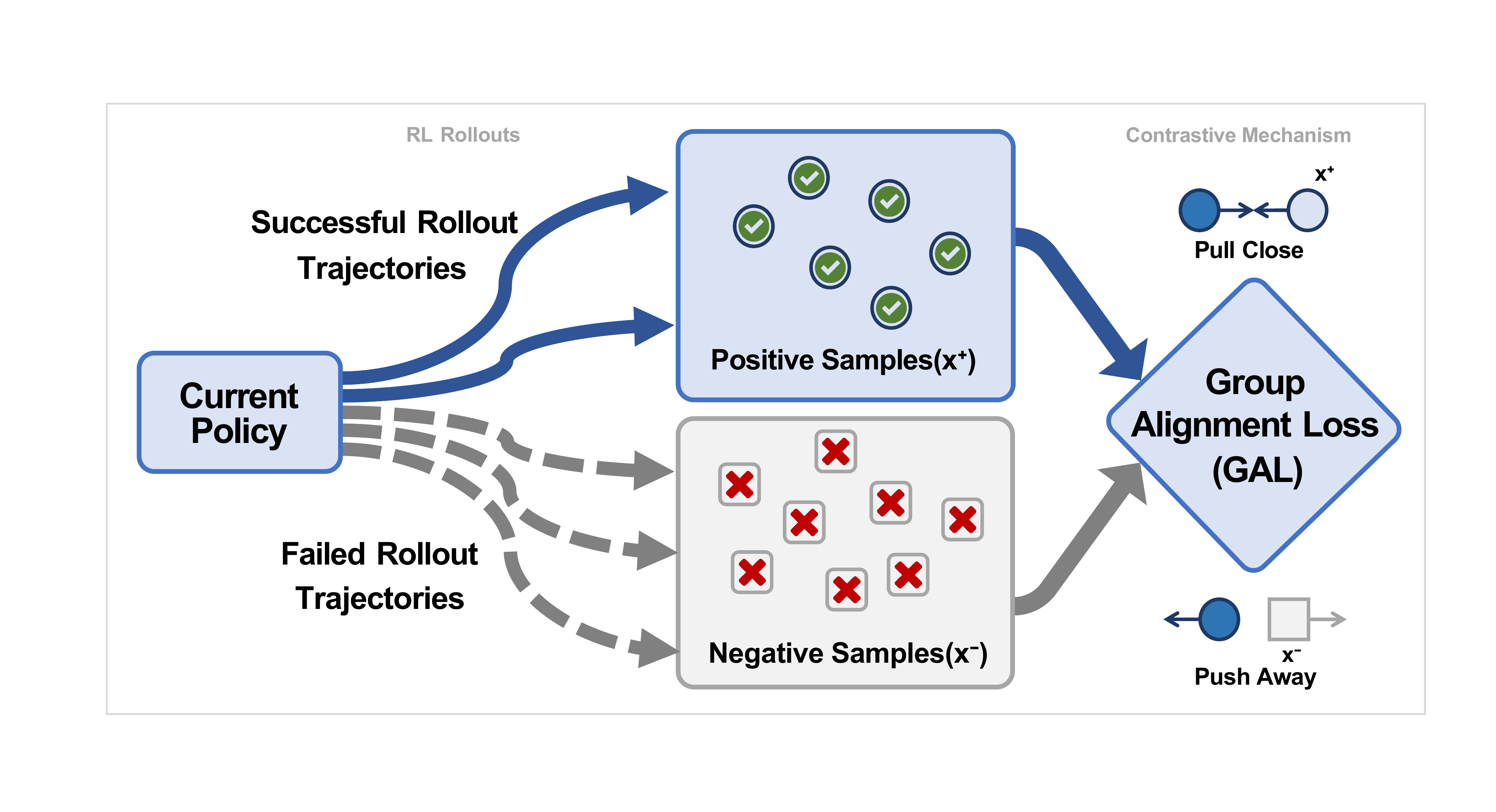}
  \caption{The Contrastive Mechanism in GAL.}
  \label{fig:GAL}
\end{figure}

\paragraph{Group Alignment Loss (GAL).}
To mitigate this, we introduce GAL. As illustrated in Figure~\ref{fig:GAL}, the current policy first generates a group of rollouts, which are categorized into positive samples (successful trajectories) and negative samples (failed trajectories) based on correctness. The core intuition is to explicitly widen the gap between these two groups by ``pulling'' the policy towards correct reasoning paths while ``pushing'' it away from incorrect ones. 

Although GAL adopts a DPO-shaped contrastive form, it is not standard offline DPO. In DYPO, GAL is constructed from \emph{on-policy} rollout groups sampled from the current policy, and its role is to serve as a variance-control term for GRPO rather than to align the model to a static preference dataset. Formally, we implement this by minimizing the following pairwise contrastive loss:

\begin{equation}
\label{eq:gal_loss}
    \mathcal{L}_{\text{GAL}}(\theta) = \mathbb{E}_{\substack{\tau_s, \tau_f \in G \\ R(\tau_s) > R(\tau_f)}} \bigg[ -\log \sigma \big( \beta_{\text{GAL}} \cdot d(\tau_s, \tau_f) \big) \bigg]
\end{equation}
where $\beta_{\text{GAL}}$ is an inverse-temperature coefficient controlling the contrastive margin, and $d(\tau_s, \tau_f)$ represents the log-ratio difference between the successful trajectory $\tau_s$ and the failed trajectory $\tau_f$, defined as:

\begin{equation}
    d(\tau_s, \tau_f) = \log \frac{\pi_\theta(\tau_s|q)}{\pi_{\text{ref}}(\tau_s|q)} - \log \frac{\pi_\theta(\tau_f|q)}{\pi_{\text{ref}}(\tau_f|q)}
\end{equation}

By applying the chain rule, the gradient of GAL is:
\begin{equation}
\begin{split}
    g_{\text{GAL}} &= -\beta_{\text{GAL}} \underbrace{(1 - \sigma(\beta_{\text{GAL}} d))}_{\text{bounded weight } w_d} \\[-5pt]
    &\quad \cdot (\nabla_\theta \log \pi_s - \nabla_\theta \log \pi_f)
\end{split}
\end{equation}
Unlike the unbounded $\hat{A}_i$ in GRPO, the weighting term $w_d$ is strictly bounded in $(0, 1)$. Let $\eta = \mathbb{E}[(1-\sigma)^2]$ represent the \textit{discrimination difficulty}. The variance of GAL (averaged over $M$ pairs) is:
\begin{equation}
    \text{Var}(g_{\text{GAL}}) \approx \frac{2 \beta_{\text{GAL}}^2 \eta}{M} \Sigma_s
\end{equation}
As the model learns to distinguish correct paths, $\sigma \to 1$ and $\eta \to 0$, causing $\text{Var}(g_{\text{GAL}}) \to 0$. Thus, GAL acts as a gradient variance reducer.

In the RL regime, we combine these objectives using a mixing coefficient $\alpha \in (0, 1)$:
\begin{equation}
    g_{\text{mix}} = \alpha g_{\text{GRPO}} + (1-\alpha) g_{\text{GAL}}
\end{equation}
Assuming independence between the exploration noise of GRPO and the discrimination noise of GAL, the variance of the combined gradient is:
\begin{equation}
\begin{aligned}
    \text{Var}(g_{\text{mix}}) &\approx \alpha^2 \text{Var}(g_{\text{GRPO}}) + (1-\alpha)^2 \text{Var}(g_{\text{GAL}}) \\
    &= \alpha^2 \left( \frac{\Sigma_s}{k} \right) + (1-\alpha)^2 \left( \frac{2\beta_{\text{GAL}}^2 \eta \Sigma_s}{M} \right)
\end{aligned}
\end{equation}
Since $\alpha < 1$ and $\eta \to 0$, it strictly follows that $\text{Var}(g_{\text{mix}}) < \text{Var}(g_{\text{GRPO}})$.

\paragraph{Summary.}
Analytically, we establish that the combined objective strictly bounds the gradient variance compared to GRPO (i.e., $\text{Var}(g_{\text{mix}}) < \text{Var}(g_{\text{GRPO}})$). Crucially, this stabilization is dynamic: as the model distinguishes successful trajectories $\tau_s$ from failed ones $\tau_f$, the discrimination difficulty $\eta$ decays to zero, naturally annealing the variance of GAL. This identifies GAL not merely as an auxiliary loss, but as an adaptive regularizer that actively dampens the high-variance noise of RL exploration.

\section{Experiments}

\subsection{Setup}
\label{sec:exp_setup}

\begin{table*}[t]
\centering
\setlength{\tabcolsep}{1pt} 

\resizebox{\textwidth}{!}{
\begin{tabular}{lcccccc|ccc}
\toprule
\multirow{2}{*}{\textbf{Model}} & \multicolumn{6}{c}{\textbf{In-Distribution}} & \multicolumn{3}{c}{\textbf{Out-of-Distribution}} \\
\cmidrule(lr){2-7} \cmidrule(lr){8-10}
 & \textbf{AIME 24} & \textbf{AIME 25} & \textbf{AMC} & \textbf{MATH-500} & \textbf{Minerva} & \textbf{Avg} & \textbf{ARC-c} & \textbf{GPQA-D} & \textbf{Avg} \\
\midrule

Qwen2.5-Math-7B & 11.5 & 4.9 & 31.3 & 43.6 & 7.4 & 19.7 & 18.2 & 11.1 & 14.6 \\

\midrule
\rowcolor{gray!10} \multicolumn{10}{c}{\textbf{Supervised Fine-Tuning}} \\
\midrule

\hspace{1em} SFT & 22.2 & 22.3 & 52.8 & 82.6 & \underline{40.8} & 44.1 & 75.2 & 24.7 & 50.0 \\

\midrule
\rowcolor{gray!10} \multicolumn{10}{c}{\textbf{Reinforcement Learning}} \\
\midrule
\hspace{1em} RL & 25.1 & 15.3 & 62.0 & 84.4 & 39.3 & 45.2 & \textbf{82.3} & 40.4 & \underline{61.4} \\
\hspace{1em} SimpleRL-Zero & 27.0 & 6.8 & 54.9 & 76.0 & 25.0 & 37.9 & 30.2 & 23.2 & 26.7 \\
\hspace{1em} OpenReasoner-Zero & 16.5 & 15.0 & 52.1 & 82.4 & 33.1 & 39.8 & 66.2 & 29.8 & 48.0 \\
\hspace{1em} PRIME-Zero & 17.0 & 12.8 & 54.0 & 81.4 & 39.0 & 40.8 & 73.3 & 18.2 & 45.8 \\
\hspace{1em} Oat-Zero & \underline{33.4} & 11.9 & 61.2 & 78.0 & 34.6 & 43.8 & 70.1 & 23.7 & 46.9 \\

% --- Section: SFT and RL ---
\midrule
\rowcolor{gray!10} \multicolumn{10}{c}{\textbf{SFT and RL}} \\
\midrule
\hspace{1em} SFT $\to$ RL & 25.8 & 23.1 & 62.7 & 87.2 & 39.7 & 47.7 & 72.4 & 24.2 & 48.3 \\
\hspace{1em} SuperRL & 28.1 & 21.6 & 63.9 & 86.4 & 36.4 & 47.3 & 77.8 & 36.9 & 57.4 \\
\hspace{1em} LUFFY & 29.4 & 23.1 & 65.6 & 87.6 & 37.5 & 48.6 & 80.5 & 39.9 & 60.2 \\
\hspace{1em} ReLIFT & 28.3 & 22.9 & 65.1 & 87.4 & 37.1 & 48.2 & 74.9 & \underline{40.9} & 57.9 \\
\hspace{1em} SRFT & 30.7 & \underline{26.0} & \textbf{69.8} & 88.4 & 39.7 & \underline{50.9} & 81.6 & 40.4 & 61.0 \\
\hspace{1em} CHORD & 31.2 & 24.4 & 66.8 & \textbf{89.4} & 39.3 & 50.2 & 81.1 & 40.4 & 60.8 \\

% --- Main Results (DYPO) ---
\midrule
\rowcolor{blue!10} \hspace{1em} \textbf{DYPO} & \textbf{36.0} (+10.2) & \textbf{28.7} (+5.6) & \underline{67.0} (+4.3) & \underline{89.2} (+2.0) & \textbf{42.4} (+2.7) & \textbf{52.5} (+4.8) & \underline{81.8} (+9.4) & \textbf{41.4} (+17.2) & \textbf{61.6} (+13.3) \\

\bottomrule
\end{tabular}
}
\caption{Overall performance on five competition-level mathematical reasoning benchmarks and two out-of-distribution benchmarks(Qwen2.5-Math-7B). Best results are \textbf{bolded} and second-best are \underline{underlined}.}
\label{tab:main_results}
\end{table*}

\textbf{Dataset Construction.} 
We align our data setup with LUFFY~\cite{yan2025learningreasonoffpolicyguidance}, utilizing the OpenR1-Math-220k~\cite{faceopen} subset with prompts primarily sourced from NuminaMath 1.5~\cite{numinamath}. To facilitate multi-teacher distillation, we employ DeepSeek-R1~\cite{guo2025deepseek} and Qwen3-235B-A22B~\cite{qwen3} to generate auxiliary reasoning traces. This ensemble strategy enriches the supervision signal and mitigates the policy's reliance on any single teacher's potentially biased reasoning patterns. All data will be open-sourced together with the code.

\noindent\textbf{Implementation Details.}
Experiments were executed on a computing cluster equipped with 2 nodes, each containing 8 $\times$ NVIDIA A800 GPUs (80GB memory). To ensure fairness, we generate 8 trajectories (rollouts) per prompt for all trained models, with a maximum response length of 8,192 tokens. The learning rate is fixed at $1 \times 10^{-6}$.
Our training pipeline is built upon the verl framework~\cite{sheng2024hybridflow}. 
For the inference and rollout phases, we utilize vLLM~\cite{vllm} to ensure high-throughput generation. 
All models were trained using bfloat16 precision to ensure numerical stability and efficiency.

\noindent\textbf{Benchmarks.}
Our method is evaluated on five in-distribution (ID) benchmarks, including {AIME 2024/2025}, {AMC}~\cite{aime}, {MATH-500}~\cite{dataset_math}, and {Minerva}~\cite{dataset_minerva}, as well as two out-of-distribution (OOD) tasks: {ARC-c}~\cite{arc} and {GPQA-Diamond}~\cite{gpqa}. Performance is measured using pass@32 for the AIME/AMC subsets and pass@1 for the others. All inference is conducted with a temperature of 0.6 and option shuffling to prevent data leakage.

\noindent\textbf{Baselines.}
We employ Qwen2.5-Math-7B~\cite{qwen2.5} and Qwen3-4B-Base~\cite{qwen3} as our base models and compare against four categories of baselines:
(1) \textit{Standard Supervised Baseline}, specifically the vanilla SFT;
(2) \textit{Zero-shot RL methods}, including SimpleRL-Zero~\cite{zeng2025simplerl}, OpenReasoner-Zero~\cite{orz}, PRIME-Zero~\cite{prime}, and Oat-Zero~\cite{oat};
(3) \textit{Post-SFT Optimization methods}, covering SFT $\to$ RL, LUFFY~\cite{yan2025learningreasonoffpolicyguidance}, ReLIFT~\cite{RELIFT},SRFT~\cite{fu2025srft}, SuperRL~\cite{liu2025superrlreinforcementlearningsupervision} and CHORD~\cite{zhang2025onpolicyrlmeetsoffpolicy}.

\subsection{Main Results}

\subsubsection{Performance on Reasoning Benchmarks}
As presented in Table~\ref{tab:main_results} (Qwen2.5-Math-7B) and Table~\ref{tab:qwen3_results} (Qwen3-4B-Base), DYPO demonstrates consistent superiority across varying model architectures. On the Qwen2.5 benchmark, DYPO achieves an average in-distribution score of 52.5, setting a new state-of-the-art.

\noindent\textbf{Comparison with SFT.}
DYPO significantly outperforms the SFT baseline, achieving a +8.4\% average improvement on Qwen2.5-Math-7B and a substantial +18.8\% on Qwen3-4B-Base. 

\noindent\textbf{Comparison with Zero-shot RL Methods.}
DYPO demonstrates superior stability over pure RL approaches like SimpleRL-Zero and Oat-Zero, surpassing the latter by a combined +19.4 points on the challenging AIME 24/25 benchmarks. While zero-shot methods often suffer from the high-variance "exploration trap," DYPO mitigates this by dynamically balancing the exploitation of priors with the exploration of new solutions, ensuring a more stable policy optimization process.

\noindent\textbf{Comparison with Multi-stage Pipelines.}
DYPO maintains a clear lead over complex pipelines (SuperRL, LUFFY, ReLIFT, SRFT, CHORD), notably outperforming SRFT, by +4.8 points on AIME 25. This advantage is echoed in Qwen3 results, where DYPO outstrips the SFT$\to$RL pipeline by +10.8\%. Unlike the uniform optimization strategies in standard pipelines, DYPO employs a dynamic mechanism that mitigates gradient vanishing on both trivial and extremely hard samples, thereby maximizing sample utilization. Furthermore, the integration of multi-teacher distillation and GAL ensures a robust and stable training trajectory, avoiding the collapse often seen in complex pipelines.

\subsubsection{Generalization to Out-of-Distribution Tasks}
Table~\ref{tab:main_results} demonstrates that DYPO avoids the generalization degradation typically associated with in-domain optimization, achieving a top average OOD score of 61.6. On the PhD-level GPQA-Diamond benchmark, it outperforms both the standard SFT baseline (+16.7\%). This indicates that DYPO transcends simple template memorization; by refining the reasoning policy rather than overfitting to surface-level patterns, it successfully transfers logical capabilities to diverse scientific domains.

\begin{table}[t]
\centering
\setlength{\tabcolsep}{3pt} 
\resizebox{\linewidth}{!}{
\begin{tabular}{lccccc|ccc}
\toprule
\multirow{2}{*}{\textbf{Model}} & \multicolumn{5}{c}{\textbf{In-Distribution}} & \multicolumn{3}{c}{\textbf{Out-of-Distribution}} \\
\cmidrule(lr){2-6} \cmidrule(lr){7-9}
 & \textbf{AIME 24/25} & \textbf{AMC} & \textbf{MATH-500} & \textbf{Minerva} & \textbf{Avg} & \textbf{ARC-c} & \textbf{GPQA-D} & \textbf{Avg} \\
\midrule

% --- Baselines ---
Qwen3-4B-Base & 9.3/5.3 & 40.0 & 66.8 & 27.9 & 29.9 & 49.4 & 14.1 & 31.8 \\
\midrule
\hspace{1em} SFT & 33.3/27.3 & 62.9 & 73.8 & 43.0 & 48.1 & 73.8 & 28.8 & 51.3 \\
% \midrule
\hspace{1em} RL & 40.6/37.3 & 71.8 & \underline{91.0} & \underline{46.3} & 57.4 & 76.7 & \underline{29.3} & 53.0 \\
% \midrule
\hspace{1em} SFT $\to$ RL & \underline{43.3/39.3} & \underline{75.4} & 77.4 & 44.9 & \underline{56.1} & \underline{77.4} & 27.8 & \underline{52.6} \\

\midrule
\rowcolor{blue!10} \hspace{1em} \textbf{DYPO} & \textbf{59.3/44.0} & \textbf{86.0} & \textbf{94.6} & \textbf{50.4} & \textbf{66.9} & \textbf{92.5} & \textbf{44.4} & \textbf{68.5} \\
\bottomrule
\end{tabular}
}
\caption{Overall performance on mathematical reasoning benchmarks (Qwen3-4B-Base). Best results are \textbf{bolded} and second-best are \underline{underlined}.}
\label{tab:qwen3_results}
\end{table}

\subsection{Ablation Study}
As shown in Table~\ref{tab:ablation_study}, we present an incremental analysis of the DYPO framework under different teacher strengths. The \textit{+ Multi-Teacher} variant serves as a data-matched supervised baseline, isolating the effect of stronger teacher supervision from the subsequent RL and routing components. Across all teacher settings (235B / 32B / 8B), performance improves monotonically as we add \textit{+RL}, \textit{+Dynamic Grading}, and \textit{+GAL}, showing that DYPO is not merely distillation with stronger teachers. Even with the weaker 8B teacher, DYPO improves AIME 25 from 22.0 to 27.8 and GPQA-D from 30.8 to 39.4, demonstrating that the RL and routing components contribute substantial gains beyond supervision alone. Overall, Dynamic Difficulty Grading brings the largest jump on the hardest reasoning benchmarks, while GAL further stabilizes optimization and yields the best final performance across all teacher scales.

\begin{table*}[t]
\centering
\setlength{\tabcolsep}{4pt}
\resizebox{\linewidth}{!}{
\begin{tabular}{lcccc}
\toprule
\textbf{Model} & \textbf{AIME 24} & \textbf{AIME 25} & \textbf{AMC} & \textbf{GPQA-D} \\
\midrule
Qwen2.5-Math-7B & 11.5 & 4.9 & 31.3 & 11.1 \\
+ SFT & 22.2 & 22.3 & 52.8 & 24.7 \\
+ Multi-Teacher (235B / 32B / 8B) & 26.6 / 26.8 / 24.5 & 23.3 / 23.5 / 22.0 & 61.4 / 61.8 / 59.5 & 33.3 / 32.3 / 30.8 \\
+ RL (235B / 32B / 8B) & 27.3 / 26.9 / 25.0 & 26.6 / 26.0 / 25.5 & 64.1 / 63.5 / 61.0 & 34.8 / 35.4 / 31.8 \\
+ Dynamic Grading (235B / 32B / 8B) & 33.3 / 31.5 / 31.8 & 28.7 / 27.9 / 28.1 & 63.6 / 62.0 / 62.4 & 36.4 / 35.4 / 34.3 \\
\midrule
\rowcolor{blue!10} + GAL (DYPO) (235B / 32B / 8B) & \textbf{36.0 / 35.2 / 33.5} & \textbf{28.7 / 28.5 / 27.8} & \textbf{67.0 / 66.5 / 65.2} & \textbf{41.4 / 41.4 / 39.4} \\
\bottomrule
\end{tabular}
}
\vspace{-5pt}
\caption{Ablation study under different teacher strengths.}
\label{tab:ablation_study}
\end{table*}

\subsection{Offline Data Ratio, Reward and Entropy}
To characterize the learning dynamics of DYPO, we monitor the Offline Data Ratio, Training Reward, and Policy Entropy across optimization steps.

\begin{figure*}[h]
    \centering
    \includegraphics[width=1.0\linewidth]{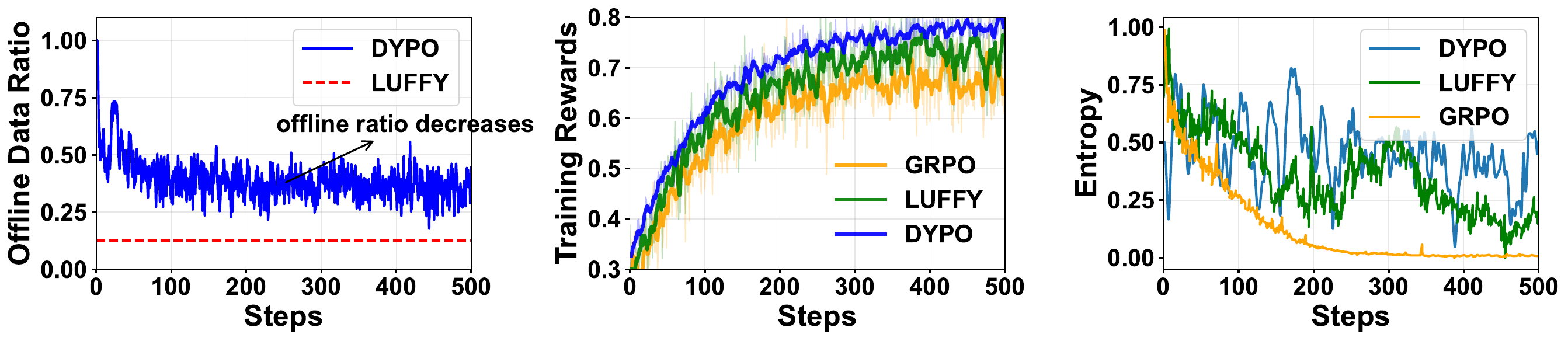} 
    \vspace{-25pt}
            \caption{\textbf{Left:} Offline data ratio over steps.\textbf{Mid:} Training reward; \textbf{Right:} Policy entropy.}
    \label{fig:training_dynamics}
\end{figure*}

Unlike static mixing (e.g., LUFFY), DYPO exhibits a self-evolving curriculum (Figure~\ref{fig:training_dynamics}, left). The Offline Data Ratio transitions from full supervision ($1.0$ at $t=0$) to a stable exploration-heavy state ($\approx 0.35$). This positioning suggests that DYPO treats offline demonstrations as a dynamic anchor: it autonomously de-leverages teacher signals as reasoning proficiency grows, yet retains a supervision floor to prevent distribution drift. The middle and right panels reveal the trade-off between convergence and diversity. While GRPO achieves rapid optimization, it suffers from premature mode collapse. 
In contrast, DYPO equilibrates reward maximization with policy stochasticity, maintaining robust entropy ($0.2 \sim 0.6$). This sustained diversity prevents the model from memorizing narrow reasoning templates, serving as the primary driver for its superior OOD generalization.

\subsection{Empirical Analysis: Gradient Stability}
\label{sec:empirical}
\begin{figure}[h]
    \centering
    \includegraphics[width=0.95\linewidth]{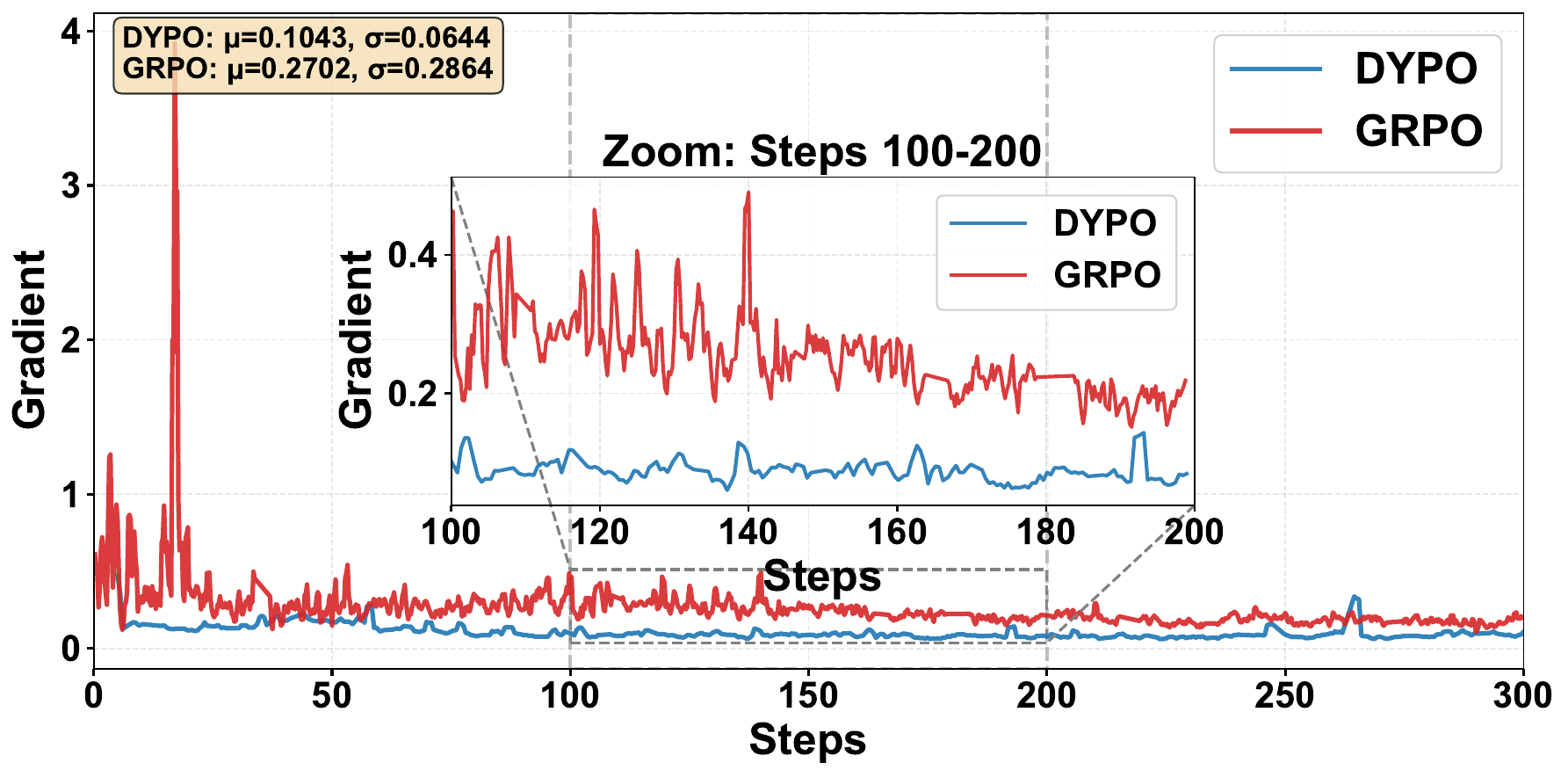}
    \vspace{-10pt}
    \caption{Gradient Norm Comparison.}
    \label{fig:grad_norm}
\end{figure}

A core theoretical contribution of DYPO is structurally resolving the bias-variance trade-off. We validate this empirically by analyzing the gradient norms of the policy network.
As shown in Figure~\ref{fig:grad_norm}, standard GRPO (red) suffers from extreme volatility, implying a rugged landscape that complicates convergence. In contrast, DYPO (blue) maintains a significantly smoother trajectory. These results confirm that our offline component acts as an effective control variate, smoothing gradient estimates to allow for more aggressive learning rates. 

\section{Related Work}

\subsection{Post-Training Paradigms for LLM Reasoning}
Enhancing the reasoning capabilities of LLMs has shifted from inference-time guidance (e.g., CoT prompting~\cite{wei2022chain}) to robust post-training strategies~\cite{wang2023aligninglargelanguagemodels}. 
Current mainstream paradigms generally fall into two categories: SFT and RL. 
SFT effectively injects knowledge and stabilizes training by fitting high-quality demonstrations~\cite{sanh2022multitask, wei2021finetuned}, yet it suffers from high fitting bias and limited OOD generalization due to its reliance on static templates~\cite{lv2025towards}. 
Conversely, RL-based methods (e.g., RLHF or RLVR) encourage models to explore the reasoning space and maximize rewards, significantly boosting performance on complex tasks~\cite{ouyang2022training, guo2025deepseek, team2025kimi}. 
However, RL gradients are inherently high-variance and unstable, particularly when valid reward signals are sparse for weaker base models~\cite{ramamurthy2022reinforcement, shao2024deepseekmath}. 
To combine these strengths, the traditional "SFT-then-RL" pipeline~\cite{touvron2023llama, yoshihara2025practical} is widely adopted but incurs multi-stage computational overhead and risks propagating SFT-induced biases into the exploration phase~\cite{lv2025towards}.

\subsection{Unified Training and Optimization Trade-offs}
To overcome the limitations of sequential pipelines, recent research focuses on unifying SFT and RL into a single-stage optimization process. 
Early attempts utilized simple loss weighting or fixed coefficients to balance stability and exploration~\cite{fu2025srft, yan2025learningreasonoffpolicyguidance}. 
More advanced approaches employ dynamic scheduling or dual-control mechanisms (e.g., CHORD, HPT) to adjust the contribution of on-policy and off-policy data during training~\cite{zhang2025onpolicyrlmeetsoffpolicy}. 
While recent theoretical works have explored the unified view of these objectives~\cite{lv2025towards}, a rigorous formalization of the gradient-level bias-variance trade-off in SFT-RL fusion remains underexplored.
Existing unified methods largely operate at a "surface level" by re-weighting scalar losses, failing to structurally resolve the statistical conflict between the high-bias SFT vector and the high-variance RL vector. 
Unlike these approaches, our DYPO framework efficiently harmonizes SFT and RL via dynamic difficulty grading, while structurally reducing RL variance through GAL and mitigating SFT bias via multi-teacher distillation.
\section{Conclusion}

We address the inherent conflict between SFT fitting bias and RL gradient variance through DYPO, a unified framework that structurally mitigates this trade-off. Unlike static weighting approaches, DYPO employs Dynamic Difficulty Grading to adaptively route queries, leveraging Multi-Teacher Distillation to correct supervisory bias and GAL to suppress gradient variance. 

Theoretically, we substantiate that DYPO achieves a linear rate of bias reduction while maintaining optimal variance control. Extensive empirical evaluations reveal that DYPO surpasses existing baselines, particularly in out-of-distribution scenarios. Beyond performance gains, the framework exhibits exceptional sample efficiency and architectural adaptability, proving effective across a spectrum of modern open-weights models. Ultimately, by dynamically governing the interplay between exploration and exploitation, DYPO establishes a unified and scalable paradigm for the next generation of reasoning-enhanced LLMs.

\clearpage

\section{Limitations}

Despite the promising performance of DYPO, we acknowledge certain limitations in our current study. First, our evaluation is primarily concentrated on logic-intensive domains, specifically mathematical reasoning tasks. While DYPO demonstrates superior stability in these objective-driven contexts, its efficacy on open-ended scenarios, such as creative writing or general chit-chat, remains to be fully explored. Second, regarding training efficiency, our method requires generating 8 trajectories per prompt to ensure robust dynamic estimation and fair comparison. This extensive online sampling inevitably introduces higher computational overhead and lower sample efficiency compared to offline baselines, representing a trade-off between optimization stability and training cost.

% Bibliography entries for the entire Anthology, followed by custom entries
%\bibliography{anthology,custom}
% Custom bibliography entries only
\bibliography{1}

\appendix
\section{Mathematical Derivations and Theoretical Analysis}
\label{sec:appendix_math}

This appendix presents the rigorous mathematical formulations of the proposed loss functions and provides a theoretical analysis of their properties, specifically focusing on gradient variance comparisons and bias reduction mechanisms within the DYPO framework.

\subsection{Loss Function Definitions}
\label{sec:loss_definitions}

\subsubsection{Conditional SFT Loss (Hard Regime)}
\label{sec:sft_loss}
In the \textbf{Hard} regime, where the policy fails to generate valid signals (i.e., all generated trajectories receive zero reward), exploration becomes inefficient. The framework strictly applies supervised fine-tuning using a Multi-Teacher strategy:

\begin{equation}
\label{eq:sft_multi}
    \mathcal{L}_{\text{SFT}}(\theta) = \mathbb{E}_{\tau_{\text{tgt}} \sim \mathcal{U}(\{\tau^{(1)}, \dots, \tau^{(m)}\})} \left[ -\log \pi_{\theta}(\tau_{\text{tgt}}|q) \right]
\end{equation}
where $\{\tau^{(1)}, \dots, \tau^{(m)}\}$ are candidate solutions generated by $m$ distinct teacher models, and $\mathcal{U}$ denotes the uniform distribution over these candidates.

\subsubsection{GRPO Loss}
\label{sec:grpo_loss}
The Group Relative Policy Optimization (GRPO) loss optimizes the policy via reinforcement learning by leveraging relative feedback within a group of sampled trajectories. Formally, for a query $q \sim \mathcal{D}$ (where $\mathcal{D}$ denotes the training data distribution), we sample a group of $k$ trajectories $G = \{\tau_1, \dots, \tau_k\}$ from the current policy $\pi_{\theta}$. The objective is defined as:

\begin{equation}
\begin{gathered}
    \mathcal{L}_{\text{GRPO}}(\theta) = -\mathbb{E}_{q \sim \mathcal{D}} \Bigg[ \frac{1}{k} \sum_{i=1}^{k} \Big( \text{CLIP}(\rho_i, \hat{A}_i, \epsilon) \Big) \\
    - \beta_{\text{KL}} \mathbb{D}_{\text{KL}}(\pi_\theta || \pi_{\text{ref}}) \Bigg]
\end{gathered}
\end{equation}

\noindent where $\rho_i(\theta) = \frac{\pi_{\theta}(\tau_i \mid q)}{\pi_{\text{ref}}(\tau_i \mid q)}$ denotes the probability ratio between the current policy and the reference policy $\pi_{\text{ref}}$. The function $\text{CLIP}(\cdot)$ represents the standard clipping mechanism with hyperparameter $\epsilon$ to constrain policy updates. The term $\mathbb{D}_{\text{KL}}$ denotes the Kullback-Leibler divergence used to prevent mode collapse, weighted by the coefficient $\beta_{\text{KL}}$.

The advantage function $\hat{A}(\tau_i, G)$ is computed using group-based standardization:
\begin{equation}
\hat{A}(\tau_i, G) = \frac{R(\tau_i, q) - \mu_G}{\sigma_G + \xi}
\end{equation}
where $\mu_G$ and $\sigma_G$ represent the mean and standard deviation of rewards within group $G$, respectively, and $\xi$ is a small constant added for numerical stability.

\subsubsection{Group Alignment Loss (GAL)}
\label{sec:gal_loss}
To mitigate the high variance associated with pure RL in the \textbf{Mid} regime, we employ a contrastive objective. We leverage the intrinsic quality differences within the sampled group $G$ by constructing pairwise comparisons. For a tuple $(q, \tau_s, \tau_f)$ drawn from $G$ where $R(\tau_s) > R(\tau_f)$ (implying $R(\tau_s)=1$ and $R(\tau_f)=0$ in binary settings):

\begin{equation}
\label{eq:gal_loss_appendix}
\begin{aligned}
\mathcal{L}_{\text{GAL}}(\theta) = \mathbb{E}_{\substack{\tau_s, \tau_f \in G \\ R(\tau_s) > R(\tau_f)}} \bigg[ -\log \sigma \bigg( \beta_{\text{GAL}} \cdot \\
\qquad \bigg( \log \frac{\pi_\theta(\tau_s|q)}{\pi_{\text{ref}}(\tau_s|q)} - \log \frac{\pi_\theta(\tau_f|q)}{\pi_{\text{ref}}(\tau_f|q)} \bigg) \bigg) \bigg]
\end{aligned}
\end{equation}
Here, $\sigma(\cdot)$ denotes the sigmoid function, and $\beta_{\text{GAL}}$ is the inverse temperature parameter controlling the discrimination margin.

\subsubsection{Unified Objective and Difficulty Grading}
\label{sec:unified_loss}
The core of DYPO is the dynamic dispatching of queries based on the rollout outcome $G$. We define three mutually exclusive indicator functions based on the set of rewards $\{R(\tau) | \tau \in G\}$:

\begin{itemize}[leftmargin=*, topsep=2pt, itemsep=0pt, parsep=0pt]
    \item \textbf{Indicator for Easy ($\mathbb{I}_{\mathcal{E}}$):} $\mathbb{I}(\forall \tau \in G, R(\tau)=1)$. The loss is set to 0 to discard trivial samples.
    \item \textbf{Indicator for Hard ($\mathbb{I}_{\mathcal{H}}$):} $\mathbb{I}(\forall \tau \in G, R(\tau)=0)$. This triggers the SFT fallback.
    \item \textbf{Indicator for Mid ($\mathbb{I}_{\mathcal{M}}$):} $\mathbb{I}(\exists \tau_i, \tau_j \in G, R(\tau_i) \neq R(\tau_j))$. This triggers the variance-reduced RL.
\end{itemize}

The final unified training objective is formulated as:
\begin{equation}
\begin{aligned}
\mathcal{L}_{\text{DYPO}}(\theta) &= \mathbb{E}_{q \sim \mathcal{D}} \bigg[ \underbrace{\mathbb{I}_{\mathcal{H}} \cdot \gamma \mathcal{L}_{\text{SFT}}}_{\text{Hard}} \\
&\quad + \underbrace{\mathbb{I}_{\mathcal{M}} \cdot \big( \alpha \mathcal{L}_{\text{GRPO}} + (1-\alpha) \mathcal{L}_{\text{GAL}} \big)}_{\text{Mid}} \bigg]
\end{aligned}
\end{equation}
Note that when $\mathbb{I}_{\mathcal{E}}=1$, the gradient contribution is effectively zero, implementing the sample discarding strategy.

\noindent\textbf{Hyperparameters:}
\begin{itemize}[leftmargin=*, topsep=2pt, itemsep=0pt, parsep=0pt]
    \item $\alpha \in [0, 1]$: Weighting coefficient balancing the summation-based RL (GRPO) and contrastive alignment (GAL).
    \item $\gamma > 0$: Scaling factor for the supervised loss component.
\end{itemize}

\subsection{Detailed Derivation of Multi-Teacher Bias Reduction}
\label{sec:appendix_multi_teacher_proof}

In this section, we provide the formal derivation for the bias reduction property of the Multi-Teacher Distillation strategy discussed in Section \ref{sec:method_hard}. We base our analysis on the bias decomposition formulation provided in the main text.

\subsubsection{Definitions and Assumptions}

Let $\tau^* \in \mathbb{R}^d$ be the optimal reasoning path (ground truth). The reasoning path generated by the $i$-th teacher, $\tau^{(i)}$, is modeled as:
\begin{equation}
    \tau^{(i)} = \tau^* + \mathbf{b}_{\text{sys}} + \mathbf{b}_i
\end{equation}
where $\mathbf{b}_{\text{sys}}$ is the systematic bias and $\mathbf{b}_i$ is the idiosyncratic bias.

To facilitate the derivation, we formalize the properties of the idiosyncratic bias term $\mathbf{b}_i$:

\begin{assumption}[Zero-Mean Idiosyncratic Bias]
We assume that the idiosyncratic biases from different teachers are independent and centered around zero in the semantic space. That is, for any teacher $i$:
\begin{equation}
    \mathbb{E}[\mathbf{b}_i] = \mathbf{0}
\end{equation}
\end{assumption}

\begin{assumption}[Variance Definition]
We define the magnitude of the idiosyncratic noise for a single teacher as $\bar{\sigma}_{\text{bias}}^2$. Formally, this is the expected squared Euclidean norm of the bias vector:
\begin{equation}
    \mathbb{E}[\|\mathbf{b}_i\|^2] = \bar{\sigma}_{\text{bias}}^2
\end{equation}
\end{assumption}

\subsubsection{Derivation of Squared Bias}

We compare the expected squared bias (estimation error) between the single-teacher baseline and the multi-teacher ensemble.

\paragraph{1. Single-Teacher SFT ($m=1$).}
When supervision is provided by a single randomly selected teacher $k$, the bias is simply $\text{Bias}_{\text{single}} = \tau^{(k)} - \tau^* = \mathbf{b}_{\text{sys}} + \mathbf{b}_k$. The expected squared norm is:
\begin{equation}
\begin{aligned}
    \mathbb{E}[\|\text{Bias}_{\text{single}}\|^2] 
    &= \mathbb{E}[\|\mathbf{b}_{\text{sys}} + \mathbf{b}_k\|^2] \\
    &= \|\mathbf{b}_{\text{sys}}\|^2 + \mathbb{E}[\|\mathbf{b}_k\|^2] \\
    &\quad + 2 \mathbf{b}_{\text{sys}}^\top \underbrace{\mathbb{E}[\mathbf{b}_k]}_{=\mathbf{0}} \\
    &= \|\mathbf{b}_{\text{sys}}\|^2 + \bar{\sigma}_{\text{bias}}^2
\end{aligned}
\end{equation}

\paragraph{2. Multi-Teacher SFT ($m > 1$).}
In the Multi-Teacher strategy, the effective supervision converges to the expectation over the sampled teachers, which is equivalent to the ensemble mean $\bar{\tau} = \frac{1}{m}\sum_{i=1}^m \tau^{(i)}$.
The effective bias vector is:
\begin{equation}
    \text{Bias}_{\text{multi}} = \left( \frac{1}{m}\sum_{i=1}^m \tau^{(i)} \right) - \tau^* = \mathbf{b}_{\text{sys}} + \frac{1}{m}\sum_{i=1}^m \mathbf{b}_i
\end{equation}

The expected squared norm of the multi-teacher bias is:
\begin{equation}
\label{eq:appendix_multi_deriv}
\begin{aligned}
    \mathbb{E}[\|\text{Bias}_{\text{multi}}\|^2] 
    &= \mathbb{E}\left[ \left\| \mathbf{b}_{\text{sys}} + \frac{1}{m}\sum_{i=1}^m \mathbf{b}_i \right\|^2 \right] \\
    &= \|\mathbf{b}_{\text{sys}}\|^2 + \mathbb{E}\left[ \left\| \frac{1}{m}\sum_{i=1}^m \mathbf{b}_i \right\|^2 \right] \\
    &\quad + 2 \mathbf{b}_{\text{sys}}^\top \underbrace{\mathbb{E}\left[ \frac{1}{m}\sum_{i=1}^m \mathbf{b}_i \right]}_{=\mathbf{0}}
\end{aligned}
\end{equation}

We focus on the variance term (the second term). Due to the independence of $\mathbf{b}_i$, the cross-terms $\mathbb{E}[\mathbf{b}_i^\top \mathbf{b}_j]$ for $i \neq j$ are zero. Thus:
\begin{equation}
\begin{aligned}
    \mathbb{E}\left[ \left\| \frac{1}{m}\sum_{i=1}^m \mathbf{b}_i \right\|^2 \right] &= \frac{1}{m^2} \sum_{i=1}^m \mathbb{E}[\|\mathbf{b}_i\|^2] \\
    &= \frac{1}{m^2} \cdot m \cdot \bar{\sigma}_{\text{bias}}^2 \\
    &= \frac{\bar{\sigma}_{\text{bias}}^2}{m}
\end{aligned}
\end{equation}

Substituting this back into Eq. \eqref{eq:appendix_multi_deriv}, we obtain the final expression presented in the main text:
\begin{equation}
    \mathbb{E}[\|\text{Bias}_{\text{multi}}\|^2] = \|\mathbf{b}_{\text{sys}}\|^2 + \frac{\bar{\sigma}_{\text{bias}}^2}{m}
\end{equation}

\subsubsection{Conclusion}

By comparing the two results, we formally establish the reduction inequality:
\begin{equation}
    \underbrace{\|\mathbf{b}_{\text{sys}}\|^2 + \frac{\bar{\sigma}_{\text{bias}}^2}{m}}_{\mathbb{E}[\|\text{Bias}_{\text{multi}}\|^2]} < \underbrace{\|\mathbf{b}_{\text{sys}}\|^2 + \bar{\sigma}_{\text{bias}}^2}_{\mathbb{E}[\|\text{Bias}_{\text{single}}\|^2]}
\end{equation}
This confirms that increasing the ensemble size $m$ strictly reduces the stochastic component of the supervisory bias.

\subsection{Gradient Variance Analysis}
\label{sec:gradient_variance}

We analyze the variance of the gradient estimators to theoretically justify the stability properties of the DYPO framework. We define the scalar variance of a gradient estimator $g$ as $\text{Var}(g) = \mathbb{E}[\|g - \mathbb{E}[g]\|^2]$.

\subsubsection{Variance of GRPO Loss}
\label{sec:grpo_variance}
Consider the gradient of the GRPO loss for a single query $q$ with group size $k$. In practical optimization, gradients are averaged over the group. The gradient is defined as:
\begin{equation}
g_{\text{GRPO}} = \frac{1}{k} \sum_{i=1}^k \nabla_\theta \log \pi_\theta(\tau_i|q) \cdot \hat{A}_i
\end{equation}
Let $s_i = \nabla_\theta \log \pi_\theta(\tau_i|q)$ be the score function. Assuming sample independence and normalized advantages ($\mathbb{E}[\hat{A}_i^2] \approx 1$), the variance is:
\begin{equation}
\label{eq:var_grpo}
\begin{aligned}
\text{Var}(g_{\text{GRPO}}) &= \frac{1}{k^2} \sum_{i=1}^k \mathbb{E}[\|s_i\|^2] \cdot \mathbb{E}[\hat{A}_i^2] \\
&\approx \frac{1}{k^2} \cdot (k \cdot \Sigma_s) = \frac{\Sigma_s}{k}
\end{aligned}
\end{equation}
where $\Sigma_s = \mathbb{E}\left[ \left\| \nabla_\theta \log \pi_\theta(\tau_i|q) \right\|^2 \right]$. This shows $\text{Var}(g_{\text{GRPO}}) \propto 1/k$.

\subsubsection{Variance of Group Alignment Loss}
\label{sec:gal_variance}
For the GAL objective, we construct $M$ preference pairs. The gradient is averaged over these pairs:
\begin{equation}
g_{\text{GAL}} = \frac{1}{M} \sum_{j=1}^M (1 - \sigma(d_j)) \cdot \beta_{\text{GAL}} (s_{s,j} - s_{f,j})
\end{equation}
Assuming independence between pairs, the variance is bounded by:
\begin{equation}
\label{eq:var_gal}
\text{Var}(g_{\text{GAL}}) \approx \frac{2 \beta_{\text{GAL}}^2 \eta \Sigma_s}{M}
\end{equation}
where $\eta = \mathbb{E}[(1 - \sigma(d))^2]$ represents the discrimination difficulty.

\subsubsection{Variance of the Combined Gradient}
\label{sec:combined_variance}
In the RL regime (Mid), we combine these objectives using a mixing coefficient $\alpha \in (0,1)$:
\begin{equation}
    g_{\text{mix}} = \alpha g_{\text{GRPO}} + (1 - \alpha) g_{\text{GAL}}
\end{equation}

Assuming independence between the exploration noise of GRPO and the discrimination noise of GAL, the variance of the combined gradient is:
\begin{equation}
\begin{aligned}
\text{Var}(g_{\text{mix}}) &\approx \alpha^2 \text{Var}(g_{\text{GRPO}}) + (1 - \alpha)^2 \text{Var}(g_{\text{GAL}}) \\
&= \alpha^2 \left( \frac{\Sigma_s}{k} \right) + (1 - \alpha)^2 \left( \frac{2\beta^2_{\text{GAL}}\eta\Sigma_s}{M} \right)
\end{aligned}
\end{equation}

We observe that as the policy improves, the discrimination task becomes easier, causing $\eta \to 0$ (since $\sigma(d) \to 1$). Furthermore, since $\alpha < 1$ implies $\alpha^2 < 1$, the contribution of the GRPO term is strictly reduced. Therefore, under the condition that $\eta$ is sufficiently small, it strictly follows that:
\begin{equation}
\text{Var}(g_{\text{mix}}) < \text{Var}(g_{\text{GRPO}})
\end{equation}
This inequality proves that the mixed objective yields a more stable gradient estimator than using GRPO alone, facilitating smoother convergence.

\section{Qualitative Analysis and Case Studies}
\label{sec:appendix_cases}

In this section, we provide a qualitative analysis of our pipeline. We first illustrate our data construction strategy, which leverages the complementary strengths of multiple teacher models. We then present specific case studies for the SFT stage and the RL stage to demonstrate how \textsc{Dypo} enhances mathematical reasoning.

\subsection{Data Construction: Leveraging Diversity}
\label{subsec:data_construction}

Our data construction method creates a high-quality, diverse dataset by distilling reasoning capabilities from multiple teacher models (e.g., DeepSeek-R1, Qwen3-235B-A22B). As illustrated in the example below, different teachers may approach the same problem via distinct but valid reasoning paths (e.g., algebraic vs. geometric). This diversity prevents the student model from overfitting to a single reasoning pattern and improves generalization.

\noindent 
\begin{minipage}{\linewidth} 
    % Problem Box
    \begin{tcolorbox}[
        enhanced,
        title={\small \faList\ \textbf{Problem Statement}}, 
        colframe=black!75,
        colback=white,
        coltitle=white,
        attach boxed title to top left={xshift=5mm, yshift=-2mm},
        boxed title style={colback=black!75, rounded corners},
        arc=1.5mm,
        left=2mm, right=2mm, top=2mm, bottom=1mm,
        boxrule=0.5mm,
    ]
        \small
        \textbf{Question:} Given a rectangular billiard table with sides 1 and $\sqrt{2}$. A ball is shot from one of its corners at an angle of $45^{\circ}$. Will it ever fall into a pocket?
    \end{tcolorbox}
    
    \vspace{-2mm}
    \begin{center}
        \textcolor{maincolor}{\faArrowDown} 
    \end{center}
    \vspace{-2mm}

    % Ours: Stacked Layout
    \begin{tcolorbox}[
        enhanced,
        title={\small \faLayerGroup\ \textbf{Our Data Construction Method}}, 
        colframe=maincolor,
        colback=white,
        coltitle=white,
        fonttitle=\bfseries,
        arc=1.5mm,
        boxrule=0.5mm
    ]
    
        % Teacher A
        \begin{tcolorbox}[
            enhanced,
            colback=teacherAcolor, 
            frame hidden,                
            borderline west={3mm}{0pt}{maincolor!60}, 
            sharp corners=west,          
            left=4mm,                    
            title={\scriptsize \faUser\ \textbf{Teacher A (DeepSeek-R1)} \hfill \textcolor{black!50}{\sf [Baseline]}},
            coltitle=maincolor!80!black,
            attach boxed title to top left={xshift=4mm, yshift=-3mm}, 
            boxed title style={frame hidden, colback=teacherAcolor},
            top=0mm, bottom=1mm
        ]
            \small
            \textbf{Reasoning:} 
            The problem reduces to finding integers $m, n$ such that $m \cdot 1 = n \cdot \sqrt{2}$.
            This implies $\frac{m}{n} = \sqrt{2}$. Since $\sqrt{2}$ is irrational, no such integers exist.
            
            \vspace{2pt}
            \hfill \textbf{Answer:} \boxed{\text{No}}
        \end{tcolorbox}

        % Connector
        \vspace{-3mm}
        \begin{center}
            \begin{tikzpicture}
                \node[circle, fill=maincolor, text=white, inner sep=2.5pt, font=\scriptsize] {\faPlus};
            \end{tikzpicture}
        \end{center}
        \vspace{-3mm}

        % Teacher B
        \begin{tcolorbox}[
            enhanced,
            colback=teacherBcolor, 
            frame hidden,
            borderline west={3mm}{0pt}{qwenpurple!60},
            sharp corners=west,
            left=4mm,
            title={\scriptsize \faUserGraduate\ \textbf{Teacher B (Qwen3-235B)} \hfill \textcolor{qwenpurple}{\sf [Complementary]}},
            coltitle=qwenpurple!80!black,
            attach boxed title to top left={xshift=4mm, yshift=-3mm},
            boxed title style={frame hidden, colback=teacherBcolor},
            top=0mm, bottom=1mm
        ]
            \small
            \textbf{Reasoning Process:}
            \textit{1. Unfolding:} Mirroring implies trajectory line $y=x$.
            \textit{2. Commensurability:} Ratio $1 : \sqrt{2}$ is irrational.
            \textit{3. Conclusion:} Never hits a vertex.
            
            \vspace{2pt}
            \hfill \textbf{Answer:} \boxed{\text{No}}
        \end{tcolorbox}

    \end{tcolorbox}
\end{minipage}

\subsection{Case Study I: Multi-Teacher Distillation (SFT)}
\label{subsec:sft_case}

Following the data construction methodology described above, the SFT stage trains the student model to recognize multiple solution paths. Here, we visualize the reasoning process for a constrained optimization problem. As shown below, \textbf{Teacher A} employs the method of Lagrange Multipliers (calculus-based), while \textbf{Teacher B} utilizes a coordinate transformation approach (algebraic/geometric). The student model benefits from exposure to both high-quality trajectories.

% Prompt
\begin{promptbox}
Find the greatest value $M$ that the expression $7 x+10 y+z$ can take when $x, y, z$ are real numbers satisfying $x^{2}+2 x+\frac{1}{5} y^{2}+7 z^{2}=6$. In which cases is this value achieved?
\end{promptbox}

% Trajectory 1: Teacher A
\begin{trajbox}[Trajectory 1: Teacher A (Lagrange Multipliers)]{traj1}
    \begin{thinkbox}
        Okay, so I need to find the maximum value of the expression $7x + 10y + z$ given the constraint...
        Let's check the constraint equation $x^2 + 2x + (1/5)y^2 + 7z^2 = 6$.
        Using Lagrange multipliers:
        \[ \nabla f = \lambda \nabla g \]
        Which gives us the system:
        1) $7 = \lambda (2x + 2)$, 
        2) $10 = \lambda (2/5 y)$, 
        3) $1 = \lambda (14z)$
        ... [Middle reasoning omitted for brevity] ...
        Therefore, the maximum value is 55.
    \end{thinkbox}

    \vspace{0.5em}
    \par\noindent\textbf{Final Response:}
    \par
    To find the greatest value $M$, we apply the method of Lagrange multipliers...
    The maximum value is achieved when:
    \[ x = -\frac{13}{62}, \quad y = \frac{175}{31}, \quad z = \frac{1}{62} \]
    The value is $\boxed{55}$.
\end{trajbox}

% Trajectory 2: Teacher B
\begin{trajbox}[Trajectory 2: Teacher B (Coordinate Transformation)]{traj2}
    \begin{thinkbox}
        Using Cauchy-Schwarz inequality...
        Let $u = x+1$, $v = y/\sqrt{5}$, $w = z\sqrt{7}$.
        The constraint becomes $u^2 + v^2 + w^2 = 7$.
        ...
    \end{thinkbox}

    \vspace{0.5em}
    \par\noindent\textbf{Final Response:}
    \par
    The solution is derived via coordinate transformation... Result is $\boxed{55}$.
\end{trajbox}

\subsection{Case Study II: Group Alignment via RL}
\label{subsec:rl_case}

While SFT establishes fundamental reasoning, the RL stage enables the exploration of complex solution paths. To ensure stability, the \textbf{Group Alignment Loss} reduces gradient variance by contrasting trajectories within a sampled group. We demonstrate this using a functional equation problem. Here, the Group Alignment Loss reinforces the rigorous application of the chain rule (Trajectory $\tau_1$) while suppressing heuristic failures (Trajectory $\tau_2$).
% Prompt
\begin{promptbox}
For arbitrary integers $n$, find the continuous function $f(x)$ which satisfies the following equation:
\[ \lim_{h\rightarrow 0}\frac{1}{h}\int_{x-nh}^{x+nh}f(t) dt=2f(nx) \]
Note that $x$ can range over all real numbers and $f(1)=1$.
\end{promptbox}

\vspace{0.5em}

% Group Summary Table
\noindent\textbf{Group Rollout Summary ($K=8$):}
\begin{center}
\renewcommand{\arraystretch}{1.2} 
\begin{tabular}{c|l|c|c}
\hline
\textbf{ID} & \textbf{Final Answer} & \textbf{Reward} & \textbf{Status} \\
\hline
$\tau_1$ & \boxed{f(x) = x} & \textbf{1.0} & \textcolor{green!60!black}{\textbf{Successful}} \\
$\tau_2$ & \boxed{f(x) = 1} & 0.0 & Failed \\
$\tau_3$ & \boxed{f(x) = x} & \textbf{1.0} & \textcolor{green!60!black}{\textbf{Successful}} \\
$\tau_4$ & \boxed{f(x) = nx} & 0.0 & Failed \\
$\tau_5$ & \boxed{f(x) = x} & \textbf{1.0} & \textcolor{green!60!black}{\textbf{Successful}} \\
$\tau_6$ & \boxed{f(x) = 0} & 0.0 & Failed \\
$\tau_7$ & \boxed{f(x) = 1} & 0.0 & Failed \\
$\tau_8$ & \boxed{f(x) = x^n} & 0.0 & Failed \\
\hline
\end{tabular}
\end{center}

\vspace{0.5em}

% Successful Trajectory
\begin{trajbox}[Successful Trajectory ($\tau_1$): Correct Chain Rule Application]{traj-green}
    \begin{thinkbox}
        Let the given equation be:
        \[ L = \lim_{h\rightarrow 0}\frac{1}{h}\int_{x-nh}^{x+nh}f(t) dt \]
        This limit represents the derivative of the integral function with respect to $h$ at $h=0$.
        Let $F(t)$ be the antiderivative of $f(t)$. Then the integral is $F(x+nh) - F(x-nh)$.
        ...
        \[ = f(x+nh) \cdot n - f(x-nh) \cdot (-n) \]
        As $h \to 0$, this becomes:
        \[ f(x) \cdot n - f(x) \cdot (-n) = 2nf(x) \]
        ...
        By continuity, $f(x) = x$ for all real $x$.
    \end{thinkbox}

    \vspace{0.5em}
    \par\noindent\textbf{Final Response:}
    \par
    The function is $f(x) = x$. The answer is $\boxed{f(x) = x}$.
\end{trajbox}

\vspace{1em}

% Failed Trajectory
\begin{trajbox}[Failed Trajectory ($\tau_2$): Missing Boundary Derivative]{traj-red}
    \begin{thinkbox}
        We need to evaluate the limit on the LHS...
        Let's assume the standard result where $\frac{1}{2h}\int_{x-h}^{x+h}f(t)dt \to f(x)$.
        ...
        Wait, let me reconsider. If I define $G(h) = \int_{x-nh}^{x+nh} f(t) dt$.
        $G'(h) = f(x+nh) - f(x-nh)$ (using standard Fundamental Theorem).
        At $h=0$, $G'(0) = 0$.
        This implies $f(x) = 0$... or $f(x) = 1$.
    \end{thinkbox}

    \vspace{0.5em}
    \par\noindent\textbf{Final Response:}
    \par
    The function satisfies $f(x) = 1$. The answer is $\boxed{f(x) = 1}$.
\end{trajbox}

\vspace{1em}
\noindent \textit{Analysis: The Failed Trajectory ($\tau_2$) attempts to derive the limit but fails to apply the chain rule to the integration bounds ($G'(h) = f(x+nh)\cdot n - f(x-nh)\cdot (-n)$), leading to an erroneous constant solution.}
\section{Additional Experimental Analysis}
\label{app:additional_exp}

\subsection{Single-Teacher vs.~Multi-Teacher Ablation}
\label{app:single_multi_teacher}

To isolate the effect of supervision diversity, we compare the full DYPO framework using a single teacher against the same framework using two teachers, while keeping all other components unchanged. As shown in Table~\ref{tab:single_multi_teacher}, multi-teacher supervision consistently improves both in-distribution and out-of-distribution performance.

\begin{table}[h]
    \centering
    \small
    \setlength{\tabcolsep}{8pt}
    \begin{tabular}{lccc}
        \toprule
        Method & AIME 24 & AMC & GPQA-D \\
        \midrule
        DYPO (Single-Teacher) & 32.8 & 64.2 & 37.5 \\
        DYPO (2-Teacher) & \textbf{36.0} & \textbf{67.0} & \textbf{41.4} \\
        \bottomrule
    \end{tabular}
    \caption{Comparison between single-teacher and multi-teacher supervision in DYPO.}
    \label{tab:single_multi_teacher}
\end{table}

This result supports the role of multi-teacher distillation in reducing teacher-specific bias and providing a stronger prior for subsequent RL optimization.

\subsection{Hyperparameter Sensitivity and Statistical Significance}
\label{app:sensitivity_significance}

We evaluate the sensitivity of DYPO to two key hyperparameters in the Mid regime: the mixing coefficient $\alpha$ and the inverse-temperature coefficient $\beta_{\text{GAL}}$. Results in Table~\ref{tab:sensitivity} show that DYPO remains stable across a broad range of settings, with the default configuration $(\alpha=0.5, \beta_{\text{GAL}}=1)$ achieving the best overall performance.

\begin{table}[h]
    \centering
    \small
    \setlength{\tabcolsep}{7pt}
    \begin{tabular}{cccc}
        \toprule
        $\alpha$ & $\beta_{\text{GAL}}$ & AIME 24 & MATH-500 \\
        \midrule
        0.2 & 0.1 & 33.5 $\pm$ 1.4 & 87.6 $\pm$ 0.7 \\
        0.2 & 1   & 34.8 $\pm$ 0.9 & 88.5 $\pm$ 0.5 \\
        0.2 & 2   & 33.9 $\pm$ 1.2 & 87.9 $\pm$ 0.6 \\
        0.5 & 0.1 & 35.2 $\pm$ 0.8 & 88.9 $\pm$ 0.4 \\
        0.5 & 1   & \textbf{36.0 $\pm$ 1.1} & \textbf{89.2 $\pm$ 0.3} \\
        0.5 & 2   & 35.5 $\pm$ 1.0 & 88.7 $\pm$ 0.5 \\
        0.8 & 0.1 & 34.6 $\pm$ 1.3 & 88.3 $\pm$ 0.8 \\
        0.8 & 1   & 35.4 $\pm$ 1.1 & 88.8 $\pm$ 0.4 \\
        0.8 & 2   & 34.2 $\pm$ 1.5 & 88.1 $\pm$ 0.7 \\
        \bottomrule
    \end{tabular}
    \caption{Sensitivity analysis of $\alpha$ and $\beta_{\text{GAL}}$. Results are reported as mean $\pm$ standard deviation over multiple random seeds.}
    \label{tab:sensitivity}
\end{table}

We further conduct paired significance tests against strong baselines using matched random seeds. The improvements of DYPO on the main benchmarks are statistically significant ($p < 0.05$).

\section{License and Artifacts Usage}

 We utilize the Qwen models and standard reasoning datasets (e.g., AIME, MATH), which are publicly available under the Apache 2.0 or MIT licenses. Our use of these artifacts for academic research and post-training optimization is strictly consistent with their intended usage policies. We release our code and the trained DYPO model checkpoints under the MIT License to promote reproducibility. This licensing is compatible with the original access conditions of the base models and datasets used in this work.
% \section{Example Appendix}
% \label{sec:appendix}

% This is an appendix.

\end{document}